\documentclass[10pt,twocolumn,letterpaper]{article}

\usepackage[pagenumbers]{iccv} 

\usepackage{acro}
\usepackage{amsmath}
\usepackage{graphicx}
\usepackage{booktabs}
\usepackage{multirow}
\usepackage{wrapfig}
\usepackage{lipsum}

\newcommand*{\ourmodel}{3D-MOOD\@\xspace}
\newcommand*{\ourmetric}{ODS\@\xspace}

\definecolor{colorFst}{HTML}{bde6cd}
\definecolor{colorSnd}{HTML}{e4eebc}
\definecolor{colorTrd}{HTML}{fff8c5}

\input{abbrv}
\usepackage[accsupp]{axessibility}  

\definecolor{lightgray}{gray}{0.9}

\definecolor{iccvblue}{rgb}{0.21,0.49,0.74}
\usepackage[pagebackref,breaklinks,colorlinks,allcolors=iccvblue]{hyperref}

\def\paperID{3} 
\def\confName{ICCV}
\def\confYear{2025}

\title{
    3D-MOOD: Lifting 2D to 3D for Monocular Open-Set Object Detection
}

\author{
Yung-Hsu Yang\textsuperscript{1} \quad
Luigi Piccinelli\textsuperscript{1} \quad Mattia Segu\textsuperscript{1} \quad Siyuan Li\textsuperscript{1} \quad Rui Huang\textsuperscript{1,2} \\[0.15cm] \quad Yuqian Fu\textsuperscript{3} \quad Marc Pollefeys\textsuperscript{1,4} \quad  Hermann Blum\textsuperscript{1,5} \quad Zuria Bauer\textsuperscript{1}\\[0.3cm]
\small $^1$ETH Z\"urich \quad $^2$Tsinghua University \quad $^3$INSAIT, Sofia University \quad $^4$Microsoft \quad $^5$University of Bonn \\
}

\begin{document}
\maketitle
\begin{abstract}
Monocular 3D object detection is valuable for various applications such as robotics and AR/VR.
Existing methods are confined to closed-set settings, where the training and testing sets consist of the same scenes and/or object categories.
However, real-world applications often introduce new environments and novel object categories, posing a challenge to these methods.
In this paper, we address monocular 3D object detection in an open-set setting and introduce the first end-to-end \textbf{3D} \textbf{M}onocular \textbf{O}pen-set \textbf{O}bject \textbf{D}etector (\textbf{\ourmodel}). 
We propose to lift the open-set 2D detection into 3D space through our designed 3D bounding box head, enabling end-to-end joint training for both 2D and 3D tasks to yield better overall performance.
We condition the object queries with geometry prior and overcome the generalization for 3D estimation across diverse scenes.
To further improve performance, we design the canonical image space for more efficient cross-dataset training.
We evaluate \ourmodel on both closed-set settings (Omni3D) and open-set settings (Omni3D $\rightarrow$ Argoverse 2, ScanNet), and achieve new state-of-the-art results.
Code and models are available at \href{https://royyang0714.github.io/3D-MOOD}{royyang0714.github.io/3D-MOOD}.
\end{abstract}    
\section{Introduction}
\label{sec:intro}

Monocular 3D object detection (3DOD) aims to recognize and localize objects in 3D space from a single 2D image by estimating their 3D positions, dimensions, and orientations.
Unlike stereo or LiDAR-based methods, monocular 3DOD relies solely on visual cues, making it significantly more challenging yet cost-effective for robotics and AR/VR applications~\cite{Zhou2019intro, dong2022depth4robotics, wang2019depth4vehicles, park2021dd3d, he2025sequential}.

\begin{figure}[t]
    \centering
    \footnotesize
    \includegraphics[width=1.0\linewidth]{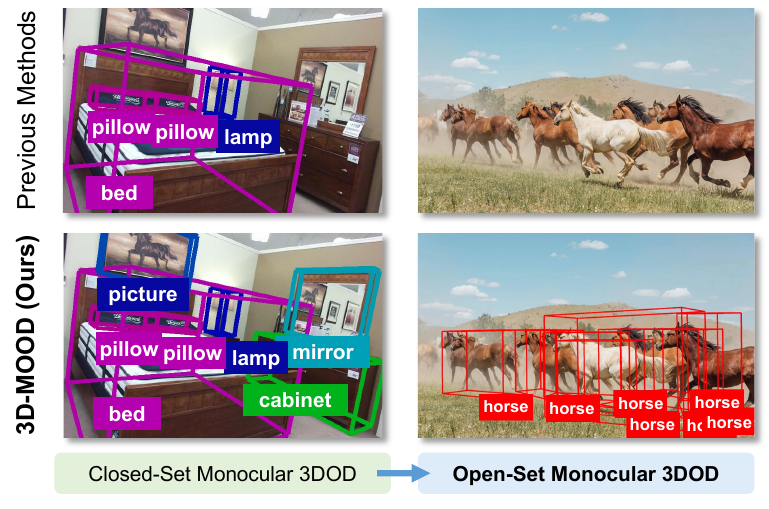}
    \vspace{-15pt}
    \caption{
        \textbf{Open-set Monocular 3D Object Detection.}
        Unlike previous methods focusing on achieving good results in the closed-set setting, we aim to resolve the open-set monocular 3D object detection problem.
        This challenge requires the model to classify arbitrary objects while precisely localizing them in unseen scenes.
    }
    \label{fig:Teaser}
    \vspace{-10pt}
\end{figure}

While many methods~\cite{wang2021fcos3d, li2022bevformer, Yang2022BEVFormerVA, liu2022petr, rukhovich2022imvoxelnet, tu2023imgeonet, zhang2022monodetr} focus on improving 3DOD performance in specific domains, Cube R-CNN~\cite{brazil2023omni3d} and Uni-MODE~\cite{Li_2024_CVPR} build unified models on the cross-dataset benchmark Omni3D~\cite{brazil2023omni3d}, which consolidates six diverse 3D detection datasets~\cite{Geiger2012CVPR,nuscenes2019,song2015sun,dehghan2021arkitscenes,objectron2021,hypersim}.
These advancements have driven the evolution of 3DOD from specialized models to more unified frameworks.
However, as shown in~\cref{fig:Teaser}, most existing methods, including the unified models, operate under an ideal assumption: the training set and testing set share identical scenes and object categories.
This limits their generalizability in real-world applications for not being able to detect novel objects in unseen domains.
This challenge motivates us \emph{to explore monocular open-set 3D object detection, further pushing the boundaries of existing 3DOD methods.}

The first step towards the open-set monocular 3DOD is identifying the fundamental obstacles underlying this task.
Our key observations are as follows: 1) \textbf{\textit{Cross-modality learning}} is crucial to breaking the limitation of closed vocabulary for novel class classification~\cite{radford2021learning}.
However, 3D data lacks rich visual-language pairs, making it significantly more challenging to learn modality alignment and achieve satisfactory open-set results.
2) \textbf{\textit{Robust depth estimation}} is essential for monocular 3DOD to generalize well across different scenes compared to LiDAR-based methods~\cite{yin2021center}.
However, monocular depth estimation particularly in novel scenes, is inherently challenging for existing methods.

Given the scarcity of 3D data and text pairs, we propose to bridge the modality gap by lifting open-set 2D detection into open-set 3D detection.
Fortunately, recent universal metric monocular depth estimation methods~\cite{piccinelli2024unidepth, piccinelli2025unidepthv2, piccinelli2025unik3d, bochkovskii2024depthpro, yin2023metric3d} have shown promising generalization across diverse scenes, which opens new opportunities for addressing open-set monocular 3DOD.
Specifically, we design a \textit{3D bounding box head} to predict the differentiable lifting parameters from 2D object queries and enable the lift of the detected 2D bounding boxes as 3D object detection.
This allows us to jointly train the open-set 2D and 3D detectors in an end-to-end (e2e) way, using both 2D and 3D ground truth (GT).
Furthermore, we propose the geometry-aware 3D query generation module, which conditions 2D object queries with the camera intrinsics and depth estimation and generates 3D object queries.
These 3D queries encode essential geometric information and are used for the 3D bounding box head to improve the model’s accuracy and generalization ability in 3D object detection.
Additionally, we design a more effective \textit{canonical image space}, which proves crucial for handling datasets with varying image resolutions, as demonstrated in our experiments.

Formally, we introduce the first e2e \textbf{3D} \textbf{M}onocular \textbf{O}pen-set \textbf{O}bject \textbf{D}etecter (\textbf{\ourmodel}) by integrating the proposed 3D bounding box head, geometry-aware 3D query generation module, and canonical image space into the open-set 2D detector~\cite{liu2023grounding}.
Our method takes a monocular input image with the language prompts and outputs the 3D object detection for the desired objects in any given scene.
Experimental results demonstrate that \ourmodel achieves state-of-the-art (SOTA) performance on the challenging closed-set Omni3D benchmark, surpassing all previous task-specific and unified models.
More importantly, in open-set settings, \ie transferring from Omni3D to Argoverse 2~\cite{Argoverse2} and ScanNet~\cite{dai2017scannet}, our method consistently outperforms prior models, achieving clear improvements in generalization and novel classes recognition.

Our main contributions are:
(1)~We explore monocular 3D object detection in open-set settings, establishing benchmarks that account for both novel scenes and unseen object categories;
(2)~We introduce \ourmodel, the first end-to-end open-set monocular 3D object detector, via 2D to 3D lifting, geometry-aware 3D query generation, and canonical image space;
(3)~We achieve state-of-the-art performance in both closed-set and open-set settings, demonstrating the effectiveness of our method and the feasibility of open-set monocular 3D object detection.

\begin{figure*}[t]
    \centering
    \footnotesize
    \includegraphics[width=1.0\linewidth]{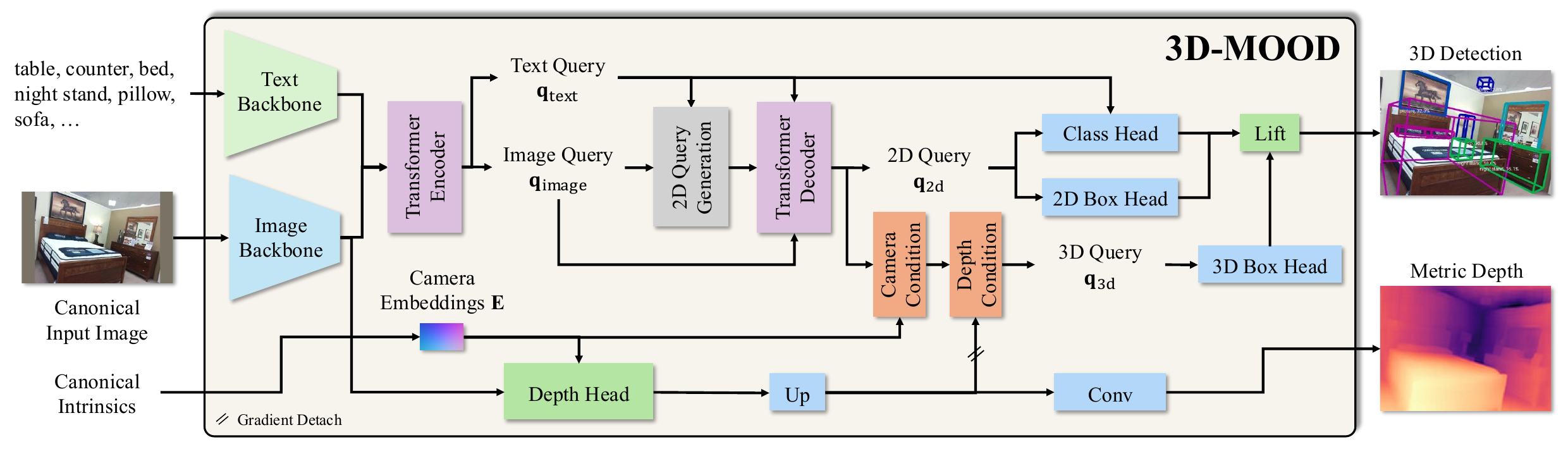}
    \vspace{-10pt}
    \caption{
        \textbf{\ourmodel.}
        We propose an end-to-end 3D monocular open-set object detector that takes a monocular image and the language prompts of the interested objects as input and classifies and localizes the 3D objects in the scenes.
        Our design will transform the input image and camera intrinsics into the proposed canonical image space and achieve the open-set ability for diverse scenes.
    }
    \label{fig:overview}
\end{figure*}

\section{Related Work}
\label{sec:related}

\subsection{Open-set 2D Object Detection}
\label{sec:related:open_2dod}
In recent years, there has been tremendous progress in 2D object detection~\cite{zareian2021open, zang2022open, liu2023grounding, zhao2024open, cheng2024yolow, fu2024cross, pan2024locate} by leveraging language models~\cite{devlin2018bert} or visual-language foundation models~\cite{radford2021learning} to detect and classify objects from language queries.
Among varying definitions of these works, \ie \textit{open-set object detection}, \textit{open-world object detection}, and \textit{open-vocabulary detection}, we do not distinguish them in this section and describe the formal problem definition in~\cref{sec:method:pre}.

OVR-CNN~\cite{zareian2021open} aligns ResNet~\cite{he2016deep} and BERT~\cite{devlin2018bert} features to detect novel classes, while OV-DETR~\cite{zang2022open} uses CLIP~\cite{radford2021learning} image and text embeddings with Detection Transformer~\cite{carion2020end}.
GLIP~\cite{liu2023grounding} presents grounded language-image pre-training to align object detection and captions.
Detic~\cite{zhou2022detecting} leverages image level labels to align object reasoning and text to enable tens of thousands of concepts for classification.

In contrast, G-DINO~\cite{liu2023grounding} deeply fuses image and text features at all stages of the detection model~\cite{zhang2022dino} and proposes language-guided query generation to allow the open-set 2D detector to detect the desired object classes according to input language prompts.
This is more natural and intuitive for humans and can help robots understand scenes in many applications.
However, in 3D monocular object detection, the lack of data annotations in 3D due to the cost also increases the difficulty of tackling the open-set classification with visual-language alignment.
Thus, in this work, we aim to propose a framework that can universally lift the open-set 2D detection to 3D to address the limitation of the data annotation for open-set classification.

\subsection{3D Monocular Object Detection}
\label{sec:related:3dod}
3D monocular object detection is crucial for autonomous driving and indoor robotic navigation.
In the past years, a large number of works~\cite{hu2022monocular,park2021dd3d,li2022bevformer,Yang2022BEVFormerVA,liu2022petr,cc3dt,lin2022sparse4d,li2022bevformer,tu2023imgeonet, naiden2019shift} proposed various methods to address the 3D multi-object detection for specific scenes, \ie one model for one dataset.
Recently, a challenging dataset called Omni3D~\cite{brazil2023omni3d} was proposed, providing a new direction for 3D monocular object detection.
This dataset contains six popular datasets including outdoor scenes (KITTI~\cite{Geiger2012CVPR} and nuScenes~\cite{nuscenes2019}), and indoor scenes (ARKitScenes~\cite{dehghan2021arkitscenes}, Hypersim~\cite{hypersim}, and SUN-RGB-D~\cite{song2015sun}), and object centric dataset (Objectron~\cite{objectron2021}).
Cube R-CNN~\cite{brazil2023omni3d} proposes virtual depth to address the various focal lengths across the diverse training datasets.
Uni-MODE~\cite{Li_2024_CVPR} proposes the domain confidence to jointly train the Bird-eye-View (BEV) detector on indoor and outdoor datasets.

Although these methods work well on Omni3D, they are still limited by the closed-set classification design, hence they lack the ability to detect novel categories.
To address this, OVM3D-Det~\cite{huang2024training} proposes a pipeline to generate pseudo GT for novel classes by using 2D foundation models~\cite{liu2023grounding, kirillov2023segany, piccinelli2024unidepth} with Large Language Model (LLM) priors.
However, while evaluating the quality of pseudo GT on open-set benchmarks, the performance is limited because the pipeline can not be e2e trained with 3D data.
On the contrary, our method is designed to estimate the differentiable lifting parameters of the open-set 2D detection with geometry prior.
Thus, it can be supervised in the e2e manner while also no longer constrained by the closed-set classification.
Furthermore, to address open-set regression in 3D, we use the canonical image space to better train 3D detectors across datasets.
With our proposed components, \ourmodel outperforms these prior works on both closed-set and open-set benchmarks.

\section{Method}
\label{sec:method}
We aim to propose the first e2e open-set monocular 3D object detector that can be generalized to different scenes and object classes.
We first discuss the problem setup in~\cref{sec:method:pre} to define the goal of monocular open-set 3D object detection.
Then, we introduce the overall pipeline of our proposed open-set monocular 3D object detector, \ourmodel, in~\cref{sec:method:gdino3d}.
We illustrate our 3D bounding box head design in~\cref{sec:method:3d_head} and introduce the proposed canonical image space for training monocular 3DOD models across datasets in~\cref{sec:method:canonical}.
In~\cref{sec:method:unidepth}, we introduce the metric monocular auxiliary depth head, which enhances 3D-MOOD by providing a more comprehensive understanding of the global scene.
Finally, in~\cref{sec:method:3d_query}, we illustrate the proposed geometry-aware 3D query generation, designed to improve generalization in both closed-set and open-set settings.

\subsection{Problem Setup}
\label{sec:method:pre}
The goal of 3D monocular open-set object detection is to detect any objects in any image, giving a language prompt for the objects of interest.
To achieve this, one needs to extend the concept of open-set beyond the distinction of seen (base) and unseen (novel)
classes within the same dataset~\cite{zareian2021open}: We follow the manner of G-DINO~\cite{liu2023grounding} that trains the model on other datasets but tests on COCO, which contains base and novel classes in unseen domains.
In this work, we aim to extend this research direction to 3DOD.
Thus, our main focus is on how to train the open-set detectors using the largest and most diverse pre-training data to date, \ie Omni3D, and achieve good performance on unseen datasets, \eg Argoverse 2 and ScanNet.

\subsection{Overall Architecture}
\label{sec:method:gdino3d}
As shown in~\cref{fig:overview}, we address the monocular open-set 3DOD by lifting the open-set 2D detection.
Formally, we estimate 2D bounding boxes $\mathbf{\hat{D}}_{\text{2D}}$ from an input image $\mathbf{I}$ and language prompts $\mathbf{T}$, and lift them as 3D orientated bounding boxes $\mathbf{\hat{D}}_{\text{3D}}$ in the corresponding camera coordinate frame with the object classes $\mathbf{\hat{C}}$.
A 2D box is defined as $\mathbf{\hat{b}}_{\text{2D}} = [\hat{x}_{1}, \hat{y}_{1}, \hat{x}_{2}, \hat{y}_{2}]$, where $\mathbf{\hat{b}}_{\text{2D}} \in \mathbf{\hat{D}}_{\text{2D}}$ in the pixel coordinate.
A 3D bounding box is defined as $\mathbf{\hat{b}}_{\text{3D}} = [\hat{x}, \hat{y}, \hat{z}, \hat{w}, \hat{l}, \hat{h}, \hat{R}]$, where $\mathbf{\hat{b}}_{\text{3D}} \in \mathbf{\hat{D}}_{\text{3D}}$.
$[\hat{x}, \hat{y}, \hat{z}]$ stands for the 3D location in the camera coordinates, $[\hat{w}, \hat{l}, \hat{h}]$ stands for the object's dimensions as width, length, and height, and $\hat{P}$ is the rotation matrix $\hat{R} \in \mathrm{SO(3)}$ of the object.

We choose G-DINO~\cite{liu2023grounding} as our 2D open-set object detector for its early visual-language features fusion design.
On top of it, we build \ourmodel with the proposed \textit{3D bounding box head}, \textit{canonical image space}, and \textit{geometry-aware 3D query generation module} for end-to-end open-set 3D object detection.
We use an image encoder~\cite{liu2021swin} to extract image features $\mathbf{q}_{\text{image}}$ from $\mathbf{I}$ and use a text backbone~\cite{devlin2018bert} from $\mathbf{T}$ to extract text features $\mathbf{q}_{\text{text}}$.
Then, following detection transformer architectures~\cite{carion2020end,zhu2020deformable,zhang2022dino}, we pass $\mathbf{q}_{\text{image}}$ and $\mathbf{q}_{\text{text}}$ to the transformer~\cite{vaswani2017attention} encoder with early visual-language features fusion~~\cite{li2022grounded}.
The image and text features will be used in the proposed language-guided query selection to generate encoder detection results $\mathbf{\hat{D}}_{\text{2D}}^{\text{enc}}$ and bounding box queries $\mathbf{q}_{\text{2d}}^{0}$ for the decoder.
For each cross-modality transformer decoder layer $\mathrm{TrD}_i$, it uses a text cross-attention $\mathrm{CA}_{\text{text}}^{i}$ and an image cross-attention $\mathrm{CA}_{\text{image}}^{i}$ to combine $\mathbf{q}_{\text{2d}}^{i}$ with the multi-modality information as:
\begin{equation}
    \begin{split}
    \mathbf{q}_{\text{2d}}^{i} &= \mathrm{CA}_{\text{text}}^{i}(\mathrm{SA}^{i}(\mathbf{q}_{\text{2d}}^{i}), \mathbf{q}_{\text{text}}), \\
    \mathbf{q}_{\text{2d}}^{i+1} &= \mathrm{FFN}^{i}(\mathrm{CA}_{\text{image}}^{i}(\mathbf{q}_{\text{2d}}^{i},\mathbf{q}_{\text{image}})),
    \end{split}
\end{equation}
where $i$ starts from 0 to $l-1$ and $\mathrm{FFN}$ stands for feed-forward neural network.
Each layer bounding box queries $\mathbf{q}_{\text{2d}}^{i}$ will be decoded as 2D bounding boxes prediction $\mathbf{\hat{D}}_{\text{2D}}^{i}$ by the 2D box head as $\mathbf{\hat{D}}_{\text{2D}}^{i} = \mathrm{MLP}_{\text{2D}}^{i}(\mathbf{q}_{\text{2d}}^{i})$, where $\mathrm{MLP}$ stands for Multi-Layer Perceptron.
The object classes $\mathbf{\hat{C}}$ are estimated based on the similarity between $\mathbf{q}_{\text{2d}}^{i}$ and the input text embeddings.

\subsection{3D Bounding Box Head}
\label{sec:method:3d_head}
Given the estimated 2D bounding boxes $\mathbf{\hat{D}}_{\text{2D}}$ and the corresponding object queries, our 3D bounding box head predict the 3D properties of $\mathbf{\hat{D}}_{\text{2D}}$ to lift it and get $\mathbf{\hat{D}}_{\text{3D}}$ in the camera coordinate frame.

\noindent{}\textbf{3D Localization.}
To localize the 3D center of the 3D bounding boxes in the camera coordinates, \ourmodel predicts the projected 3D center and the metric depth of the 3D center of the object as~\cite{hu2022monocular,cc3dt,brazil2023omni3d}.
To be more specific, we predict $[\hat{u}, \hat{v}]$ as the distance between the projected 3D center and the center of the 2D detections.
We lift the projected center to the camera coordinate with the given camera intrinsic $\mathbf{K}$ and the estimated metric depth $\hat{z}$ of the 3D bounding boxes center.
We estimate the scaled logarithmic depth prediction from our 3D bounding box head noted as $\hat{d}$ with depth scale $s_{depth}$.
Thus, the metric depth will be acquired as $\hat{z} = \exp(\hat{d} / s_{depth})$ during inference.

\noindent{}\textbf{3D Object Dimensions.}
To estimate the universal 3D objects, we follow~\cite{hu2022monocular, cc3dt} to directly predict dimensions instead of using the pre-computed category prior as in~\cite{brazil2023omni3d}.
Our bounding box head predicts the scaled logarithmic dimensions as $[s_{\text{dim}} \times \ln\hat{w}, \ln\hat{l} \times s_{\text{dim}}, ln\hat{h} \times s_{\text{dim}}]$ as the output space and can obtain the width, length, and height with exponential and divided by scale $s_{\text{dim}}$ during inference.

\noindent{}\textbf{3D Object Orientation.}
Unlike~\cite{hu2022monocular,cc3dt}, we follow~\cite{kundu20183d} to predict 6D parameterization of $\hat{R}$, denoted ad $\hat{\text{rot}}_{6d}$, instead of only estimating yaw as autonomous driving scenes.

Following detection transformer (DETR)~\cite{carion2020end}-like architecture design, we use an MLP as the 3D bounding box head to estimate the $12$ dimension 3D properties from 2D object queries $\mathbf{q}_{\text{2d}}^{i}$ for each transformer decoder layer $i$.
The 3D detection $\mathbf{\hat{D}}_{\text{3D}}^{i}$ for each layer is estimated by separate 3D bounding box heads ($\mathrm{MLP}_{\text{3D}}^{i}$) as:
\begin{equation}
\label{eq:3d_head}
    \mathbf{\hat{D}}_{\text{3D}}^{i} = \mathbf{Lift}(\mathrm{MLP}_{\text{3D}}^{i}(\mathbf{q}_{\text{2d}}^{i}), \mathbf{\hat{D}}_{\text{2D}}^{i}, \mathbf{K}).
\end{equation}
where $\mathbf{Lift}$ stands for we obtain the final 3D detections in the camera coordinate by lifting the projected 3D center with the estimated dimensions and rotation.

\begin{figure}[t]
    \small
    \footnotesize
    \centering
    \includegraphics[width=1.0\linewidth]{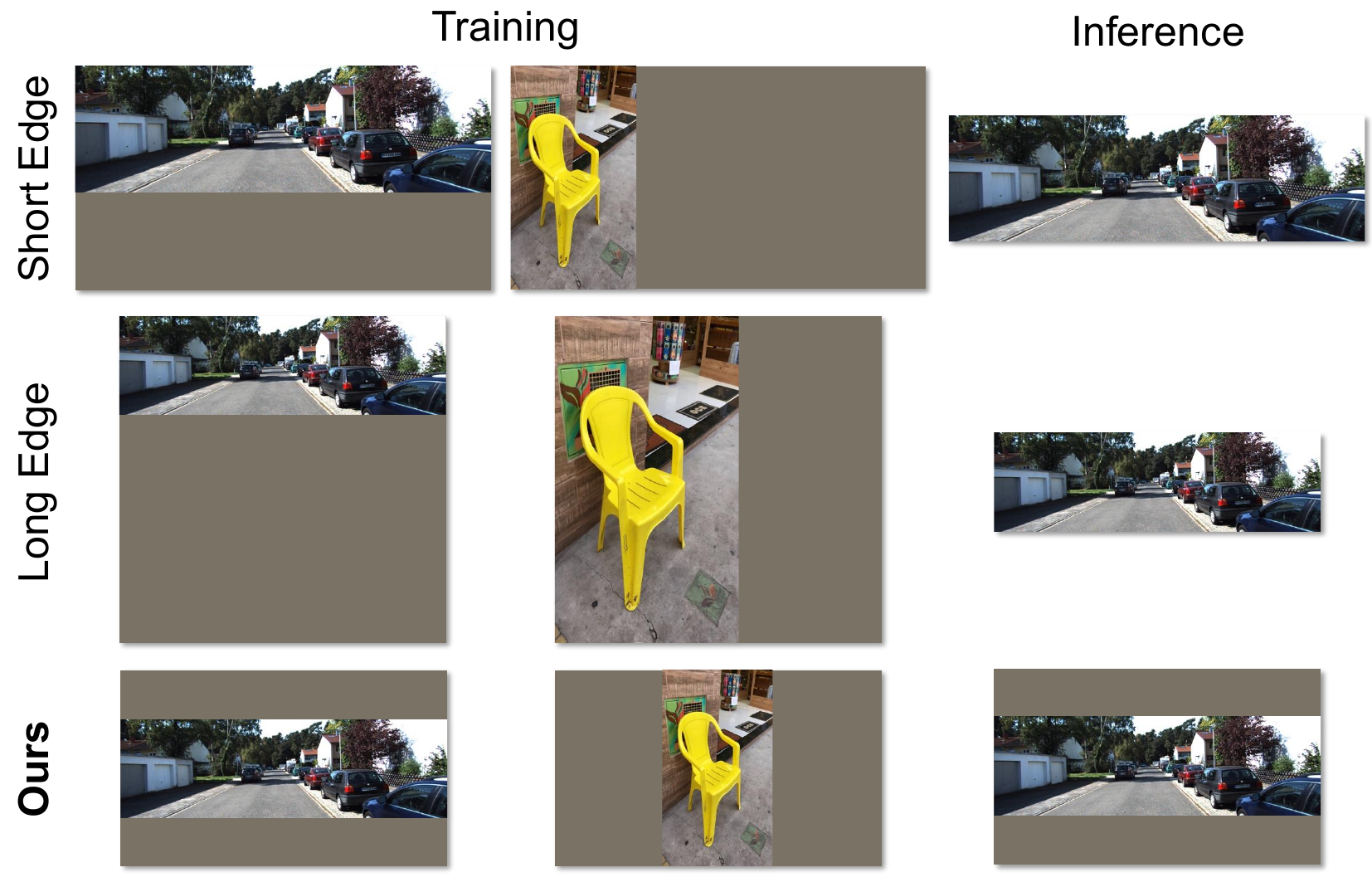}
    \vspace{-10pt}
    \caption{
        \textbf{Canonical Image Space.}
        We compare the difference between different resizing and padding strategies.
        It is worth noting that the same image will have the same camera intrinsic $\mathrm{K}$ despite having very different image resolutions for previous methods.
    }
    \label{fig:canonical}
\end{figure}

\subsection{Canonical Image Space}
\label{sec:method:canonical}
To train the model across datasets that contain images with different resolutions from various datasets, previous works~\cite{brazil2023omni3d, Li_2024_CVPR, liu2023grounding} either resize the short or long edge to a particular value, then use right and bottom padding to align the image resolution of the training batches.
However, as shown in~\cref{fig:canonical}, previous methods will heavily pad zeros when the training batches have very different resolutions and won't change the camera intrinsics.
This wastes resources for non-informative information but will also cause the same camera intrinsic $\mathbf{K}$ but with different image resolutions between training and inference time while also breaking the center projection assumption.

As illustrated in~\cite{yin2023metric3d}, the ambiguity among image, camera intrinsic, and metric depth will confuse the depth estimation model during training with multiple datasets.
Thus, we proposed the canonical space where the model can have a unified observation for both training and testing time.
We use the fixed input image resolution $[\mathbf{H}_{c} \times \mathbf{W}_{c}]$ and resize the input images and intrinsics so that the height or width reaches $\mathbf{H}_{c}$ or $\mathbf{W}_{c}$ to keep the original input image ratio.
Then, we center pad the images to $[\mathbf{H}_{c} \times \mathbf{H}_{W}]$ with value $0$ and pad the camera intrinsic accordingly.
This alignment is necessary for the model to learn the universal settings consistent across training and test time, and we demonstrate the effectiveness in closed-set and open-set experiments.
We show more details in the supplementary material.

\subsection{Auxiliary Metric Depth Estimation}
\label{sec:method:unidepth}
A significant challenge in monocular 3DOD is accurately estimating object localization in 3D.
3D object localization is directly tied to the localization in the image plane and the object metric depth, making the metric depth estimation sub-task crucial for 3DOD.
Previous methods~\cite{park2021dd3d, Yang2022BEVFormerVA, lin2023sparse4d} have emphasized the importance of incorporating an auxiliary depth estimation head to improve 3D localization.
However, achieving accurate depth localization becomes more difficult when attempting to generalize depth estimation across different datasets.
Recent methods~\cite{piccinelli2024unidepth, piccinelli2025unik3d, yin2023metric3d} demonstrate the possibility of overcoming the universal depth estimation by leveraging the camera information.
As Cube R-CNN~\cite{brazil2023omni3d} uses a similar approach as Metric3D~\cite{yin2023metric3d} to have virtual depth, we argue that conditioning depth features on camera intrinsics yields a more robust solution.
This approach avoids being limited by variations in camera models and enhances generalizability.
To this end, we design an auxiliary depth estimation head conditioned on the camera information, as proposed in UniDepth~\cite{piccinelli2024unidepth, piccinelli2025unidepthv2} to achieve a generalizable monocular depth estimation.

In particular, our model architecture incorporates an additional Feature Pyramid Network (FPN)~\cite{lin2017feature} to extract depth features $\mathbf{F}$ from the image backbone~\cite{liu2021swin}.
We rescale them to $1/16$ of the input image height $H$ and width $W$ and generate the depth features $\mathbf{F}_{16}^{d}$ using a Transformer block~\cite{vaswani2017attention}.
We condition $\mathbf{F}_{16}^{d}$ using camera embeddings, $\mathbf{E}$, as described in~\cite{piccinelli2024unidepth}.
We then upsample the depth features to $1/8$ of the input image height and width, \ie $\mathbf{F}_{8}^{d} | \mathbf{E}$ to estimate the metric depth by a convolutional block.
We generate the scaled logarithmic depth prediction $\hat{d}_\text{full}$ with the same depth scale $s_{\text{depth}}$ as our 3D bounding box head.
Thus, the final metric depth $\hat{z}_{full}$ will be acquired as $\hat{z}_{\text{full}} = \exp(\hat{d}_{\text{full}} / s_{\text{depth}})$.

\subsection{Geometry-aware 3D Query Generation}
\label{sec:method:3d_query}
To ensure the 3D bounding box estimation can be generalized for diverse scenes, we propose a geometry-aware 3D query generation to condition the 2D object query $\mathbf{q}_{\text{2d}}$ with the learned geometry prior.
First, we use the camera embeddings $\mathbf{E}$ in our auxiliary depth head to make the model aware of the scene-specific prior via a cross-attention layer.
Due to the sparsity of the 3D bounding box annotations compared to the per-pixel depth supervision, we further leverage the depth features $\mathbf{F}_{8}^{d} | \mathbf{E}$ to condition the object query.
This allows us to align the metric depth prediction and 3D bounding box estimation while leveraging the learned depth estimation.
Our full geometry-aware query generation will generate the 3D box queries $\mathbf{q}_{\text{3d}}$ as:
\begin{equation}
    \begin{split}
         \mathbf{q}_{\text{3d}}^{i} &= \mathrm{FFN}_{\text{cam}}^{i}(\mathrm{CA}_{\text{cam}}^{i}(\mathrm{SA}_{\text{cam}}^{i}(\mathbf{q}_{\text{2d}}^{i}), \mathbf{E})), \\
         \mathbf{q}_{\text{3d}}^{i} &= \mathrm{FFN}_{\text{depth}}^{i}(\mathrm{CA}_{\text{depth}}^{i}(\mathrm{SA}_{\text{depth}}^{i}(\mathbf{q}_{\text{3d}}^{i}), \mathbf{F}_{8}^{d} | \mathbf{E})).
    \end{split}
\end{equation}
We replace the 2D object queries in~\cref{eq:3d_head} with the generated 3D queries $\mathbf{q}_{\text{3d}}^{i}$ for each decoder layer as
\begin{equation}
    \label{eq:3d_query_head}
    \mathbf{\hat{D}}_{\text{3D}}^{i} = \mathbf{Lift}(\mathrm{MLP}_{\text{3D}}^{i}(\mathbf{q}_{\text{3d}}^{i}), \mathbf{\hat{D}}_{\text{2D}}^{i}, \mathbf{K}).
\end{equation}
It is worth noting that we detach the gradient from the cross attention between 3D query and depth features to stabilize the training.
We validate our geometry-aware 3D query generation in our ablation studies for both closed-set and open-set settings.
The results suggest that incorporating geometric priors enhances model convergence during closed-set multi-dataset training and improves the robustness of 3D bounding box estimation in real-world scenarios.

\subsection{Training Loss}
\label{sec:method:loss}
We train \ourmodel with 2D losses $L_\text{2D}$, 3D losses $L_\text{3D}$, and auxiliary depth loss $L_\text{depth}^\text{aux}$ in conjugation.
For 2D losses, we follow MM G-DINO~\cite{zhao2024open} and use L1 loss and GIoU~\cite{rezatofighi2019generalized} loss for the 2D bounding box regression and contrastive between predicted objects and language tokens for bounding box classification as GLIP~\cite{li2022grounded}.
For the 3D losses, we use L1 loss to supervise each estimated 3D properties.
We compute 2D and 3D losses for each transformer decoder layer $i$ and obtain $L_{\text{2D}}^i$ and $L_{\text{3D}}^i$.
For auxiliary depth estimation, we refer to each original dataset of Omni3D to find the depth GT or using the projected LiDAR points or structure-from-motion (SfM)~\cite{schoenberger2016sfm,schoenberger2016mvs} points.
We use Scale-invariant log loss~\cite{eigen2014depth} as auxiliary depth loss $L_\text{depth}^\text{aux}$ with $\lambda_{\text{depth}}$ as loss weight for supervision.
Finaly, we set the loss weights for 2D and 3D detection to $1.0$ and $\lambda_{\text{depth}}$ to $10$ and obtain the final loss $L_\text{final}$ as
\begin{equation}
    L_\text{final} = \sum_{i=0}^{l}(L_{\text{2D}}^i + L_\text{3D}^i) + \lambda_{\text{depth}}L_\text{depth}^\text{aux} .
\end{equation}

\section{Experiments}
\label{sec:exp}

We first describe our implementation details for \ourmodel and datasets in~\cref{sec:exp:implement} and discuss the evaluation metrics in~\cref{sec:exp:evaluation}.
Then, we show the open-set, cross-domain, and closed-set results in~\cref{sec:exp:open},~\cref{sec:exp:cross_domain}, and~\cref{sec:exp:omni3d}, and analyze the results of ablation studies in~\cref{sec:exp:ablation}.
We show some qualitative results in~\cref{sec:exp:qualitative} and more in the supplementary material.

\subsection{Implementation Details}
\label{sec:exp:implement}
\noindent{}\textbf{Model.}
We use the Vis4D~\cite{vis4d2024} as the framework to implement \ourmodel in PyTorch~\cite{pytorch} and CUDA~\cite{nickolls2008cuda}.
We train the full model for $120$ epochs with batch size of $128$ and set the initial learning rate of $0.0004$ following~\cite{zhao2024open}.
For the ablation studies, we train the model for $12$ epochs with batch size of $64$.
We choose $800 \times 1333$ as our canonical image shape, as described in~\cref{sec:method:canonical}.
During training, we use random resize with scales between $[0.75, 1.25]$ and random horizontal flipping with a probability of $0.5$ as data augmentation.
We decay the learning rate by a factor of $10$ at epochs $8$ and $11$ for the $12$ epoch setting and by $80$ and $110$ for the $120$ epoch setting.

\noindent{}\textbf{Closed-set Data.}
We use Omni3D~\cite{brazil2023omni3d} as training data, which contains six popular monocular 3D object detection datasets, \ie KITTI~\cite{Geiger2012CVPR}, nuScenes~\cite{nuscenes2019}, SUN RGB-D~\cite{song2015sun}, Objectron~\cite{objectron2021}, ARKitScenes~\cite{dehghan2021arkitscenes}, Hypersim~\cite{hypersim}.
There are $176573$ training images, $19127$ validation images, and $39452$ testing images with $98$ classes.
We follow~\cite{brazil2023omni3d, Li_2024_CVPR} using the training and validation split from Omni3D~\cite{brazil2023omni3d} with $50$ classes for training and test the model on the test split.

\noindent{}\textbf{Open-set Data.}
We choose two challenging datasets for indoor and outdoor scenes as the open-set monocular 3D object detection benchmarks.
For outdoor settings, we use the validation split of Argoverse 2 (AV2)~\cite{Argoverse2} Sensor Dataset as the benchmark.
We sample $4806$ images from the ring front-center camera, which provides portrait resolution ($2048 \times 1550$), and use all the official classes that appear in the validation set to be evaluated.
For indoor settings, we use the validation split of ScanNet~\cite{dai2017scannet} with official $18$ classes as the indoor benchmark.
We uniformly sample $6240$ images with $968 \times 1296$ resolution along with the axis-aligned 3D bounding boxes.
We provide more details in the supplementary material.

\subsection{Evaluation}
\label{sec:exp:evaluation}
\begin{figure}[t]
    \centering
    \footnotesize
    \includegraphics[width=1.0\linewidth]{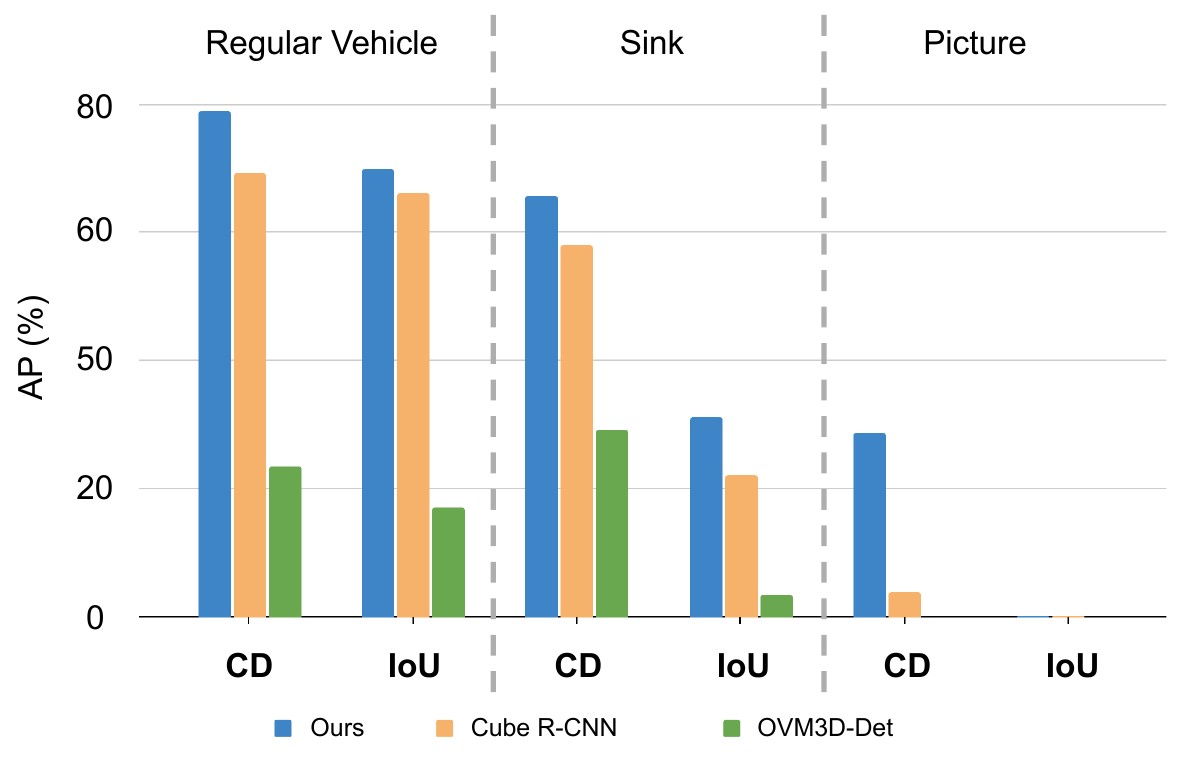}
    \vspace{-20pt}
    \caption{
        \textbf{Matching function.}
        Different matching criteria over three methods on three different classes on AV2 and ScanNet.
        CD stands for matching prediction and GT using our proposed normalized center distance matching, while IoU stands for using $\text{IoU}_\text{3D}$.
    }
    \label{fig:center_distance}
\end{figure}

We use the average precision (AP) metric to evaluate the performance of 2D and 3D detection results.
Omni3D~\cite{brazil2023omni3d} matches the predictions and GT by computing the intersection-over-union ($\text{IoU}_{\text{3D}}$) of 3D cuboids.
The mean 3D AP, \ie ${\text{AP}_\text{3D}}$, is reported across classes and over a range of $\text{IoU}_{\text{3D}}$ thresholds $ \in [0.05, 0.1, ..., 0.5]$.
However, this matching criterion is too restrictive for small or thin objects for monocular object detection, especially for open-set scenarios.
As shown in~\cref{fig:center_distance}, we report the difference between different matching criteria over three classes and methods under open-set settings.
The performance of large objects, such as Regular Vehicles (cars), remains consistent between center-distance (CD) based and 
IoU-based matching.
However, for smaller objects (\eg, Sinks) and thinner objects (\eg, Pictures), IoU-based matching fails to accurately reflect the true performance of 3D monocular object detection.
Thus, we refer to nuScenes detection score (NDS)~\cite{nuscenes2019} and composite detection score ($\mathrm{CDS}$)~\cite{Argoverse2} to propose a new 3D object detection score for open-set monocular object detection noted as open detection score (\textbf{\ourmetric}).

To use \ourmetric for both indoor and outdoor datasets, we use 3D Euclidean distance instead of the bird-eye view (BEV) distance used in autonomous driving scenes.
Furthermore, unlike NDS and CDS using the fixed distances as matching thresholds, we set the matching distances as the uniform range $ \in [0.5, 0.55, ..., 1.0]$ of the \textit{radius} of the 3D GT boxes.
This allows a flexible matching criterion given the object size and strikes a balance between IoU matching and other distance matching.
We report mean 3D AP using normalized distance-based matching as $\text{AP}_{\text{3D}}^\text{ dist}$ over classes.
We compute several true positive (TP) errors for the matched prediction and GT pair.
We report mean average translation error (mATE), mean average scale error (mASE), and mean average orientation error (mAOE) to evaluate how precise the true positive is compared to the matched GT.
The final \ourmetric is computed as the weighted sum of $\text{AP}_{\text{3D}}^\text{ dist}$, mATE, mASE, and mAOE as:
\begin{equation}
    \text{ODS} = \frac{1}{6} [3 \times \text{AP}_{\text{3D}}^\text{ dist} + \sum(1 - \text{mTPE})],
\end{equation}
where mTPE $\in$ [mATE, mASE, mAOE].
\ourmetric considers the average precision and true positive errors under the flexible distance matching, making it suitable for evaluating the monocular 3D detection results, especially for open-set settings.
In this work, we report $\text{AP}_{\text{3D}}$, $\text{AP}_{\text{3D}}^{\text{ dist}}$, and \ourmetric in the form of percentage by default.
Additional details are provided in the supplementary material.

\begin{table*}[t]
    \small
    \footnotesize
    \centering
    \caption{
        \textbf{Open-set Results}.
        We propose the first 3D monocular open-set object detection benchmark with Argoverse 2~\cite{Argoverse2} (Outdoor) and ScanNet~\cite{dai2017scannet} (Indoor).
        Each dataset contains seen (base) and unseen (novel) categories in the unseen scenes.
        Besides Cube R-CNN~\cite{brazil2023omni3d} full model, we evaluate Cube R-CNN (In/Out) as each domain expert variant, which is only trained and tested on Omni3D indoor/outdoor datasets.
        It is worth noting that OVM3D-Det's depth estimation model~\cite{piccinelli2024unidepth} is trained on AV2 and ScanNet.
        We further evaluate the generalization of seen classes and the ability to detect novel classes through \colorbox{colorSnd}{\ourmetric (B)} and \colorbox{colorTrd}{\ourmetric (N)}.
        \ourmodel establishes the SOTA performance on this new challenging open-set benchmark.
    }
    \vspace{-5pt}
    \resizebox{\linewidth}{!}{
    \setlength{\tabcolsep}{2pt}
    \begin{tabular}{l|ccccc|cc|ccccc|cc}
        \toprule
        \multirow{2}{*}{Method} & \multicolumn{7}{c|}{Argoverse 2} & \multicolumn{7}{c}{ScanNet} \\
        & $\text{AP}_{\text{3D}}^{\text{ dist}} \uparrow$ & mATE $\downarrow$ & mASE $\downarrow$ & mAOE $\downarrow$ & \ourmetric $\uparrow$ & \colorbox{colorSnd}{\ourmetric (B)} $\uparrow$ & \colorbox{colorTrd}{\ourmetric (N)} $\uparrow$ & $\text{AP}_{\text{3D}}^{\text{ dist}} \uparrow$ & mATE $\downarrow$ & mASE $\downarrow$ & mAOE $\downarrow$ & \ourmetric $\uparrow$ & \colorbox{colorSnd}{\ourmetric (B)} $\uparrow$ & \colorbox{colorTrd}{\ourmetric (N)} $\uparrow$ \\
        \midrule
        Cube R-CNN (In) & - & - & - & - & - & - & - & 19.5 & 0.725 & 0.771 & 0.858 & 20.5 & 24.6 & 0.0 \\
        Cube R-CNN (Out) & 10.5 & 0.896 & 0.869 & 0.991 & 9.3 & 19.5 & 0.0 & - & - & - & - & - & - & - \\
        Cube R-CNN~\cite{brazil2023omni3d} & 8.6 & 0.903 & 0.867 & 0.953 & 8.9 & 18.6 & 0.0 & 20.0 & 0.733 & 0.774 & 0.921 & 19.5 & 23.4 & 0.0 \\
        \midrule
        OVM3D-Det$^\dagger$~\cite{huang2024training} & 7.7 & 0.914 & 0.893 & 0.899 & 8.8 & 16.5 & 1.7 & 15.6 & 0.798 & 0.871 & 0.818 & 16.3 & 17.8 & 8.8 \\
        \midrule
        \textbf{Ours} (Swin-T) & \textbf{14.8} & 0.782 & 0.697 & 0.612 & 22.5 & 31.7 & 14.2 & 27.3 & 0.630 & 0.726 & \textbf{0.650} & 30.2 & 33.6 & 13.4 \\
        \textbf{Ours} (Swin-B) & 14.7 & \textbf{0.755} & \textbf{0.680} & \textbf{0.580} & \textbf{23.8} & \textbf{33.6} & \textbf{14.8} & \textbf{28.8} & \textbf{0.612} & \textbf{0.706} & 0.655 & \textbf{31.5} & \textbf{34.7} & \textbf{15.7} \\
        \bottomrule
    \end{tabular}}
    \label{tab:results:open-set}
    \vspace{-5pt}
\end{table*}
\begin{table*}
    \small
    \footnotesize
    \centering
    \caption{
        \textbf{Cross-domain results.}
        We validate \ourmodel cross-domain generalization by training on one of the indoor datasets from Omni3D while testing on the other two in a zero-shot manner.
        \ourmodel generalize better consistently for all three settings.
    }
    \vspace{-5pt}
    \resizebox{0.8\linewidth}{!}{
    \begin{tabular}{l|ccc|ccc|ccc}
        \toprule
        \multirow{2}{*}{Method} & \multicolumn{3}{c|}{Trained on Hypersim} & \multicolumn{3}{c|}{Trained on SUN RGB-D} & \multicolumn{3}{c}{Trained on ARKitScenes} \\
        & $\text{AP}_{\text{3D}}^{\text{ hyp}} \uparrow$ & $\text{AP}_{\text{3D}}^{\text{ sun}} \uparrow$  & $\text{AP}_{\text{3D}}^{\text{ ark}} \uparrow$  & $\text{AP}_{\text{3D}}^{\text{ hyp}} \uparrow$  & $\text{AP}_{\text{3D}}^{\text{ sun}} \uparrow$  & $\text{AP}_{\text{3D}}^{\text{ ark}} \uparrow$  & $\text{AP}_{\text{3D}}^{\text{ hyp}} \uparrow$  & $\text{AP}_{\text{3D}}^{\text{ sun}} \uparrow$  & $\text{AP}_{\text{3D}}^{\text{ ark}} \uparrow$  \\
        \midrule
        Cube R-CNN~\cite{brazil2023omni3d} & 15.2 & 9.5 & 7.5 & 9.5 & 34.7 & 14.2 & 7.5 & 13.1 & 38.6 \\
        Uni-MODE~\cite{Li_2024_CVPR} & 14.7 & 5.6 & 3.6 & 3.0 & 28.5 & 8.8 & 4.2 & 13.0 & 35.0 \\
        \textbf{Ours} & \textbf{25.6} & \textbf{15.9} & \textbf{14.5} & \textbf{13.8} & \textbf{42.1} & \textbf{21.4} & \textbf{12.9} & \textbf{23.8} & \textbf{43.9} \\
        \bottomrule
    \end{tabular}
    }
    \vspace{-10pt}
    \label{tab:rebuttal:cross_domain}
\end{table*}

\subsection{Open-Set Results}
\label{sec:exp:open}
\noindent{}\textbf{Benchmarks.}
We establish the first 3D monocular open-set object detection benchmark as~\cref{tab:results:open-set}.
We treat the diverse Omni3D~\cite{brazil2023omni3d} dataset as the training set and test the model performance on Argoverse 2 (outdoor)~\cite{Argoverse2} and ScanNet (indoor)~\cite{dai2017scannet} validation splits as open-set testing.

\noindent{}\textbf{Baselines.}
We validate the performance of \ourmodel by comparing it to several baselines.
To the best of our knowledge, there are only two methods~\cite{brazil2023omni3d, Li_2024_CVPR} trained on the entire Omni3D training set.
However, until the submission, Uni-MODE~\cite{Li_2024_CVPR} did not release their model weights.
Hence, we use Cube R-CNN~\cite{brazil2023omni3d} to build several baselines.
We further compare the generalizability of \ourmodel with Uni-MODE in~\cref{sec:exp:cross_domain}.
We use three different Cube R-CNN models, which are trained with indoor-only, outdoor-only, or full Omni3D training sets, as the specialized closed-set models for indoor (In), outdoor (Out), and universal data (Cube R-CNN).
We map the predicted categories from Omni3D to Argoverse 2 (AV2) and ScanNet to conduct the 3D detection on open data, which can provide $11$ and $15$ seen (base) classes, respectively.
Another baseline is OVM3D-Det~\cite{huang2024training}, which uses G-DINO~\cite{liu2023grounding}, SAM~\cite{kirillov2023segany}, UniDepth~\cite{piccinelli2024unidepth} and LLM to generating pseudo GT for 3D detection.
We run the OVM3D-Det pipeline on AV2 and ScanNet to generate the pseudo GT as open-set detection results and evaluate it with the real GT.

\noindent{}\textbf{Results.}
As shown in \cref{tab:results:open-set}, \ourmodel achieves the SOTA on both challenging datasets in open-set settings.
The Cube R-CNN baselines (rows 1 to 3) show that the closed-set methods lack the ability to recognize the novel objects due to the closed-vocabulary model design, which further heavily affects the overall open-set performance when more than half of classes are novel, \eg AV2.
Furthermore, the performance differences between \ourmodel and Cube R-CNN on the seen (base) classes are more significant in the unseen domain.
This suggests that \ourmodel benefits from the proposed canonical image spaces and geometry-aware 3D query generation, leading us to generalize better for unseen domains.
The comparison to OVM3D-Det~\cite{huang2024training} shows the importance of e2e design to align better 2D open-set detector and 3D object detection. 
Given that UniDepth~\cite{piccinelli2024unidepth} is trained on AV2 and ScanNet, the depth estimation from OVM3D-Det is much more accurate.
However, a lack of training in 3D data leads to worse performance for both base and novel classes.

\subsection{Cross Domain Results.}
\label{sec:exp:cross_domain}
Since we can not directly compare Uni-MODE~\cite{Li_2024_CVPR} on our proposed open-set benchmarks, we follow~\cite{brazil2023omni3d, Li_2024_CVPR} and conduct the cross-domain generalization experiments within Omni3D datasets.
We train \ourmodel on one indoor dataset at once and zero-shot test on the other two datasets.
As shown in~\cref{tab:rebuttal:cross_domain}, our method can achieve higher performance for both in-domain data, \ie seen dataset, and out-of-domain data.
We believe it demonstrates the ability of the models to detect the base object in the unseen scenes, which benefits from our geometry-aware design.

\begin{table}[t]
    \small
    \footnotesize
    \centering
    \caption{
        \textbf{Results on Omni3D.}
        We compare \ourmodel with other closed-set detectors on Omni3D test set.
        $\text{AP}^{\text{ omni}}_{\text{3D}} \uparrow $ is the average scores over Omni3D $6$ datasets.
        All methods are trained with Omni3D train and val splits and ``-" represents the results not reported in previous literature~\cite{brazil2023omni3d, Li_2024_CVPR}.
        \ourmodel achieves SOTA performance on the closed-set setting with the open-set ability.
    }
    \vspace{-3pt}
    \resizebox{\linewidth}{!}{
    \begin{tabular}{l|cccccc|c}
        \toprule
        Method & $\text{AP}^{\text{ kit}}_{\text{3D}} \uparrow$ & $\text{AP}^{\text{ nus}}_{\text{3D}} \uparrow$  & $\text{AP}^{\text{ sun}}_{\text{3D}} \uparrow$ & $\text{AP}^{\text{ hyp}}_{\text{3D}} \uparrow$ & $\text{AP}^{\text{ ark}}_{\text{3D}} \uparrow$& $\text{AP}^{\text{ obj}}_{\text{3D}} \uparrow$ & $\text{AP}^{\text{ omni}}_{\text{3D}} \uparrow$ \\
        \midrule
        ImVoxelNet~\cite{rukhovich2022imvoxelnet} & - & - & - & - & - & - & 9.4 \\
        SMOKE~\cite{liu2020smoke} & - & - & - & - & - & - & 9.6 \\
        FCOS3D~\cite{wang2021fcos3d} & - & - & - & - & - & - & 9.8 \\
        PGD~\cite{wang2022probabilistic} & - & - & - & - & - & - & 11.2 \\
        \midrule
        Cube R-CNN~\cite{brazil2023omni3d} & 32.6 & 30.1 & 15.3 & 7.5 & 41.7 & 50.8 & 23.3 \\
        Uni-MODE$^*$~\cite{Li_2024_CVPR} & 29.2 & \textbf{36.0} & 23.0 & 8.1 & 48.0 & 66.1 & 28.2 \\
        \midrule
        \textbf{Ours} (Swin-T) & \textbf{32.8} & 31.5 & 21.9 & \textbf{10.5} & 51.0 & 64.3 & 28.4 \\
        \textbf{Ours} (Swin-B) & 31.4 & 35.8 & \textbf{23.8} & 9.1 & \textbf{53.9} & \textbf{67.9} & \textbf{30.0} \\
        \bottomrule
    \end{tabular}
    }
    \vspace{-10pt}
    \label{tab:results:omni3d}
\end{table}

\begin{table*}[t]
    \small
    \footnotesize
    \centering
    \caption{
        \textbf{Ablations of \ourmodel}.
        \textbf{CI} denotes canonical image space, \textbf{Depth} denotes auxiliary depth estimation head, and \textbf{GA} stands for geometry-aware 3D query generation.
        We report the IoU-based AP for the Omni3D test split and our ODS for the AV2 and ScanNet validation split. \colorbox{colorSnd}{$\text{AP}^{\text{ omni}}_{\text{3D}}$} is the average scores over Omni3D $6$ datasets while \colorbox{colorTrd}{$\text{ODS}^\text{ open}$} is the average for open-set datasets.
        The results show that our proposed component help for both closed-set and open-set settings.
    }
    \vspace{-5pt}
    \resizebox{\linewidth}{!}{
    \begin{tabular}{cccc|ccccccc|ccc}
        \toprule
        & CI & Depth & GA & $\text{AP}^{\text{ kit}}_{\text{3D}} \uparrow $ & $\text{AP}^{\text{ nus}}_{\text{3D}} \uparrow $ & $\text{AP}^{\text{ hyp}}_{\text{3D}} \uparrow $ & $\text{AP}^{\text{ sun}}_{\text{3D}} \uparrow $ & $\text{AP}^{\text{ ark}}_{\text{3D}} \uparrow $ & $\text{AP}^{\text{ obj}}_{\text{3D}} \uparrow $ & \colorbox{colorSnd}{$\text{AP}^{\text{ omni}}_{\text{3D}}$} $\uparrow$ & $\text{ODS}^\text{ av2} \uparrow $ & $\text{ODS}^\text{ scan} \uparrow $ & \colorbox{colorTrd}{$\text{ODS}^\text{ open}$} $\uparrow$ \\
        \midrule
        1 & - & - & - & \textbf{32.5} & 29.7 & 8.1 & 17.3 & 46.5 & 54.9 & 24.1 & 18.2 & 29.0 & 23.6 \\
        2 & \checkmark & - & - & 31.1 & 30.5 & 9.1 & 19.1 & 47.7 & 58.1 & 25.5 & 19.5 & 29.5 & 24.5 \\
        3 & \checkmark & \checkmark & - & 29.8 & 30.7 & \textbf{10.3} & 19.9 & 48.6 & 58.8 & 26.2 & 20.0 & 29.4 & 24.7 \\
        4 & \checkmark & \checkmark & \checkmark & \underline{32.1} & \textbf{31.9} & \underline{9.9} & \textbf{20.8} & \textbf{49.1} & \textbf{60.2} & \textbf{26.8} & \textbf{22.0} & \textbf{30.0} & \textbf{26.0} \\
        \bottomrule
    \end{tabular}}
    \vspace{-2pt}
    \label{tab:results:ablations}
\end{table*}

\subsection{Closed-Set Results}
\label{sec:exp:omni3d}
We compare \ourmodel with the other closed-set models on the Omni3D~\cite{brazil2023omni3d} benchmark.
As shown in~\cref{tab:results:omni3d}, \ourmodel achieves the SOTA performance on Omni3D test split.
Our model with Swin-Transformer~\cite{liu2021swin} Tiny (Swin-T) as backbone achieves similar performance as previous SOTA Uni-MODE~\cite{Li_2024_CVPR}, which uses ConvNeXt~\cite{liu2022convnet} Base model.
When we use the comparable image backbone to ConvNeXt-Base (89M), \ie Swin-Transformer Base (Swin-B, 88M), \ourmodel achieves ${30.1 \%}$ AP on Omni3D test set and establish the new SOTA results on the benchmark.

\subsection{Ablation Studies}
\label{sec:exp:ablation}
We ablate each contribution in \cref{tab:results:ablations} for both closed-set and open-set settings.
We build the naive baseline as row 1 by directly using 2D queries $\mathbf{q}_{2d}$ and directly generate the 3D detection results.

\noindent{}\textbf{Canonical Image Space.}
As shown in~\cite{rezatofighi2019generalized,piccinelli2024unidepth}, it is crucial to resolve the ambiguity between image, intrinsic, and depth.
With the proposed canonical image (CI) space, we align the training and testing time image shape and camera intrinsics.
Row 2 outlines how CI improves closed-set and open-set results by $1.4$ and $0.9$ for closed-set and open-set settings, respectively.
This shows that the model learns the universal property and makes the detection ability generalize well across datasets for training and testing time.

\noindent{}\textbf{Auxiliary Depth head.}
We validate the effect of the auxiliary depth head as row 3.
Learning metric depth is essential for the network to better understand the geometry of the 3D scene instead of merely relying on the sparse depth supervision signal from the 3D bounding boxes loss.
With the auxiliary depth head, \ourmodel improves $0.7$ AP on the closed-set settings yet only slightly improve the open-set settings by $0.2$ ODS.
We hypothesize that the depth data is not diverse and rich enough in Omni3D compared to the data that other generalizable depth estimation methods~\cite{piccinelli2024unidepth,yin2023metric3d, bochkovskii2024depthpro} use for training.
Thus, the benefits from the depth head is little in open-set settings.

\noindent{}\textbf{Geometry-aware 3D Query Generation.}
Finally, we ablate our proposed geometry-aware 3D query generation module in row 4.
We show that for both closed-set and open-set settings, the geometry condition can improve the performance by $0.6$ and $1.3$, respectively.
It is worth noting that the geometry information can significantly improve the model's generalizability, which demonstrates our contribution to 3D monocular open-set object detection.

\begin{figure}[t]
    \centering
    \scriptsize
    \setlength{\tabcolsep}{1pt}
    \newcommand{\sz}{0.48}
    \newcommand{\hs}{2.4cm}
    \vspace{-5pt}
    \begin{tabular}{cc}
        \\
        \includegraphics[width=\sz\linewidth, height=\hs]{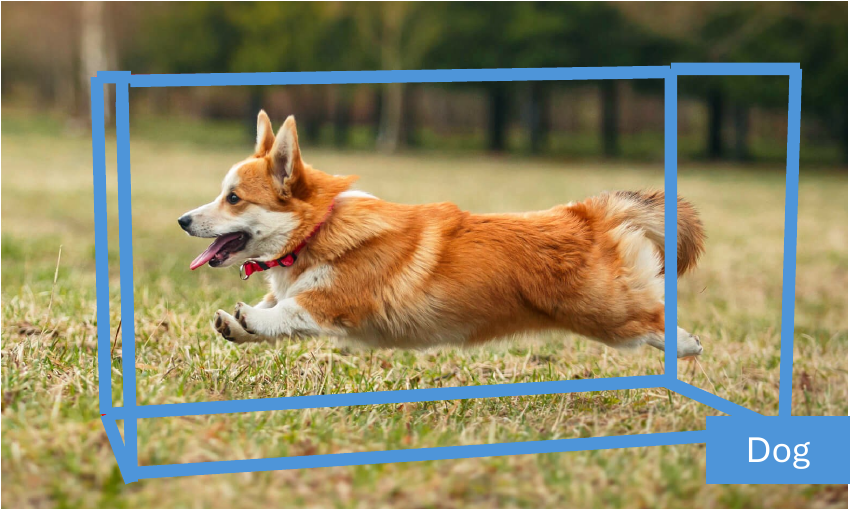} &
        \includegraphics[width=\sz\linewidth, height=\hs]{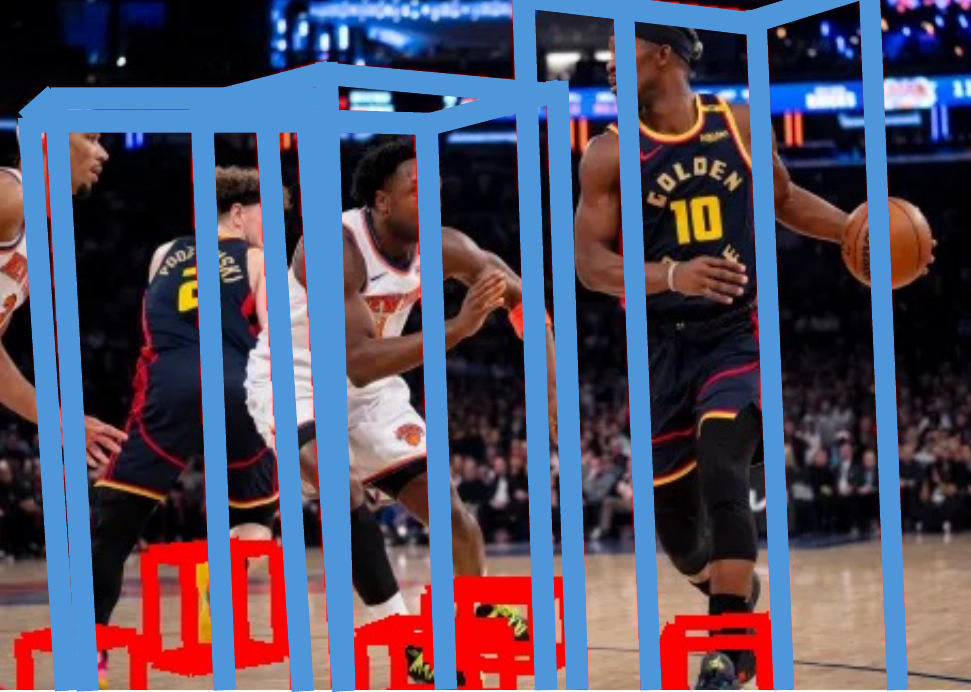} \\
        \includegraphics[width=\sz\linewidth, height=\hs]{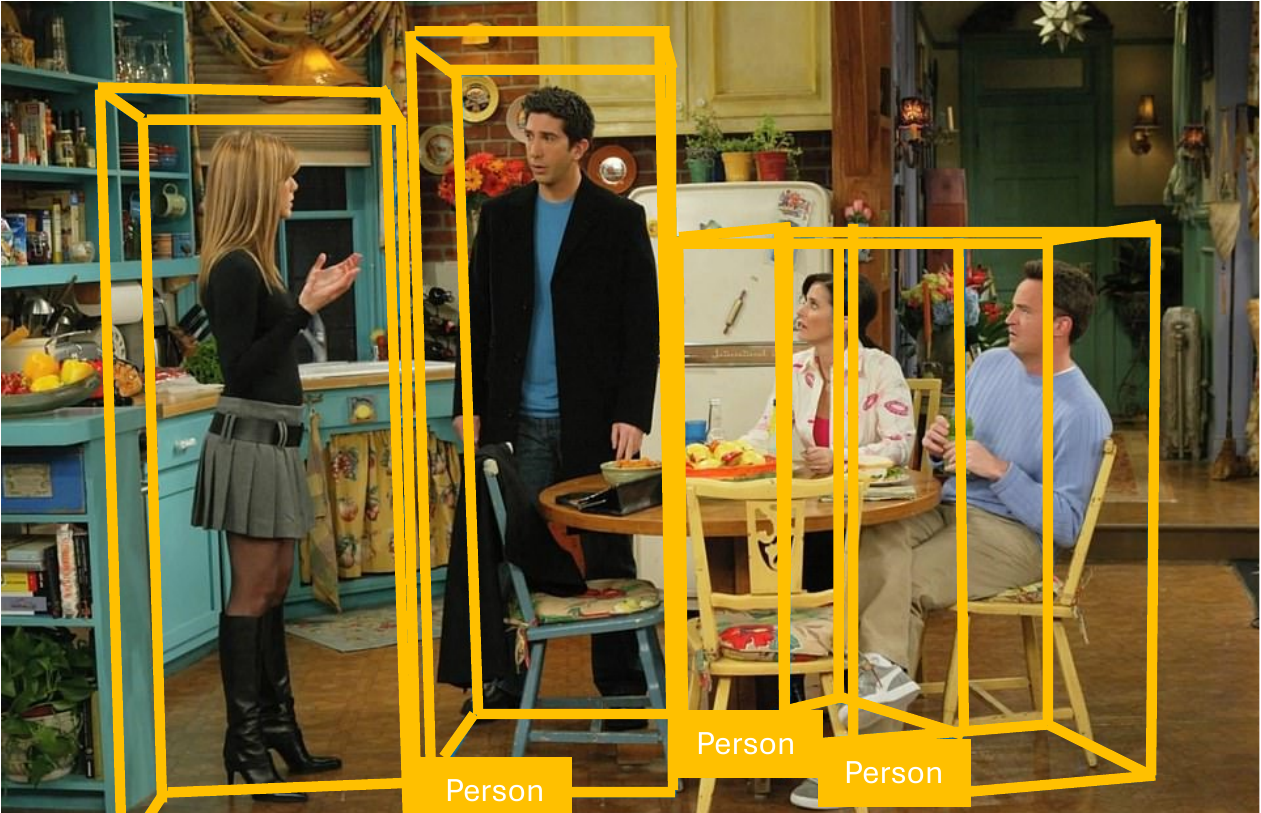} &
        \includegraphics[width=\sz\linewidth, height=\hs]{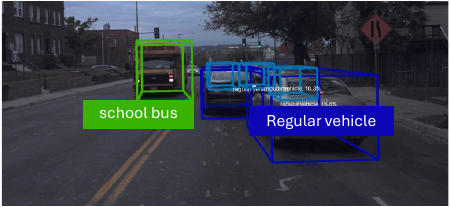} \\
        \includegraphics[width=\sz\linewidth, height=\hs]{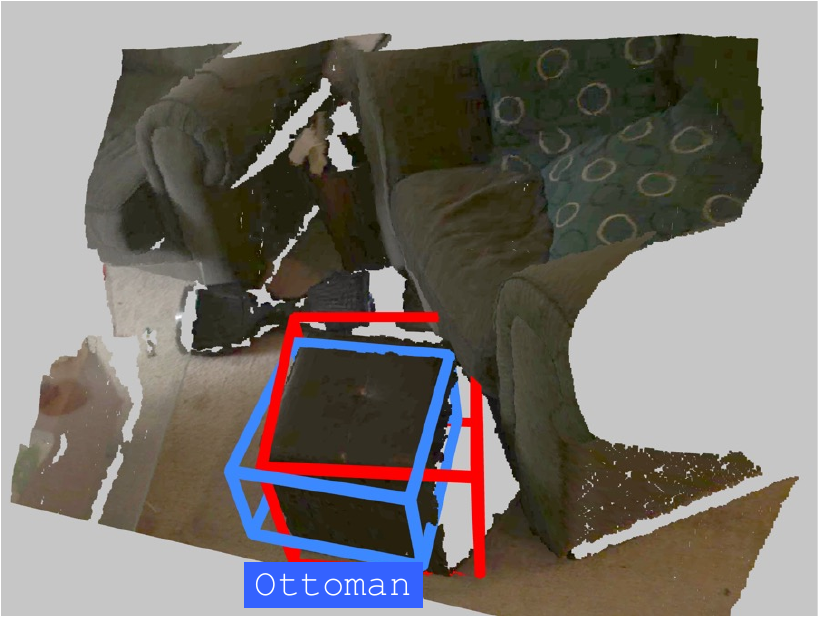} &
        \includegraphics[width=\sz\linewidth, height=\hs]{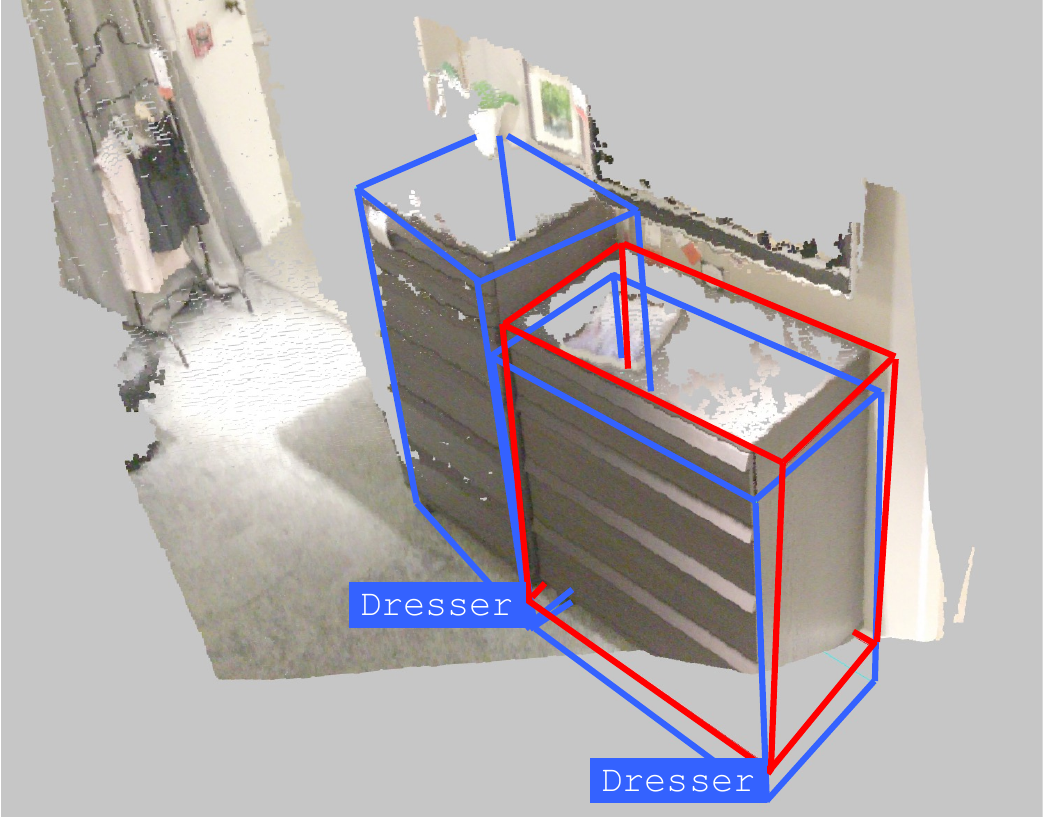} \\ 
    \end{tabular}
    \vspace{-10pt}
    \caption{
        \textbf{In-the-wild Qualitative Results.}
        We show the visualization of \ourmodel for in-the-wild images.
        The red boxes in the 3D visualization (last row) are the GT annotations.
    }
    \vspace{-15pt}
    \label{fig:qualitative_results}
\end{figure}

\subsection{Qualitative Results}
\label{sec:exp:qualitative}
We show the open-set qualitative results in \cref{fig:qualitative_results} to demonstrate the generalizability of \ourmodel, where we successfully detect novel objects in unseen scenes.
More results are reported in the supplementary material.

\section{Conclusion}
\label{sec:conclusion}
In this work, we introduce \ourmodel, the first end-to-end 3D monocular open-set object detection method, which achieves state-of-the-art performance in closed-set settings while proving strong generalization to unseen scenes and object classes in open-set scenarios.
We design a 3D bounding box head with the proposed geometry-aware 3D query generation to lift the open-set 2D detection to the corresponding 3D space.
Our proposed method can be trained end-to-end and yield better overall performance.
Furthermore, our proposed canonical image space resolves the ambiguity between image, intrinsic, and metric depth, leading to more robust results in closed-set and open-set settings.
We propose a challenging 3D monocular open-set object detection benchmark using two out-of-domain datasets.
\ourmodel sets the new state-of-the-art performance on the challenging Omni3D benchmark compared to other closed-set methods.
Moreover, the results on the open-set benchmark demonstrate our method's ability to generalize the monocular 3D object detection in the wild.

\section*{Acknowledgement}
This research is supported by the ETH Foundation Project 2025-FS-352, Swiss AI Initiative and a grant from the Swiss National Supercomputing Centre (CSCS) under project ID a03 on Alps, and the Lamarr Institute for Machine Learning and Artificial Intelligence.
The authors thank Linfei Pan and Haofei Xu for helpful discussions and technical support.

{
    \small
    \bibliographystyle{ieeenat_fullname}
    \bibliography{main}
}

\clearpage
\renewcommand\thesection{\Alph{section}}

\setcounter{section}{0}
\setcounter{page}{1}
\setcounter{table}{4}
\setcounter{figure}{5}

\maketitlesupplementary

This supplementary material elaborates more details of our main paper.
In~\cref{sec:sup:cis}, we illustrate how the proposed canonical image space can reduce the required GPU resource for training and how it helps the model learn better prior geometry.
In~\cref{sec:sup:virtual_depth}, we compare our proposed geometry-aware query generation to the virtual depth proposed by Cube R-CNN~\cite{brazil2023omni3d}.
In~\cref{sec:sup:open}, we provide further details of our proposed open-set benchmark.
We analyze the depth estimation performance in~\cref{sec:sup:mmde}, backbone comparison in~\cref{sec:sup:backbone} and FPS in~\cref{sec:sup:fps}, respectively.
In~\cref{sec:sup:eval}, we discuss the importance of our proposed open detection score (ODS) and compare the evaluation results in detail to the IoU-based AP.
Finally, we provide more qualitative results in~\cref{sec:sup:qualitative} for both closed-set and open-set settings.

\section{Canonical Image Space}
\label{sec:sup:cis}
As shown in the main paper, we compare different resizing and padding strategies for training the model.
Given the training batch size as $2$, we have two training samples having very different image ratios, \eg $[376 \times 1241]$ and $[1920 \times 1080]$.
The first strategy as~\cite{brazil2023omni3d} will find the shortest edge and resize it to the desired value, \eg $512$, and conduct the right-bottom padding to align the two samples' resolutions.
This will lead to considerable padding for the portrait image, while not change the camera intrinsic $\mathbf{K}$.

The second strategy is like Grounding DINO (G-DINO)~\cite{liu2023grounding}, which will find the longest edge and resize it to the desired value, \eg $1333$, and use the same right-bottom padding to align the resolutions.
This leads to considerable padding for the landscape image, while the padding will also not change the camera intrinsic $\mathbf{K}$.
Both strategies will increase the GPU usage for unnecessary padding and lead to different image resolutions for the same sampled image according to the paired images.

On the other hand, our proposed canonical image space fixes the image resolutions, \eg $[800 \times 1333]$, and will resize the longest or the shortest edge considering the image ratios.
As shown in \cref{tab:canonical}, our methods successfully reduce the needed GPU resources compared to the previous methods. 
Furthermore, we use the center padding to ensure our image space will affect the camera intrinsic $\mathbf{K}$ accordingly to unify it across not only the training and testing time but also across datasets.

During inference time, the same camera will capture the same image shape with the same camera intrinsics.
The previous methods will fail to align the same observation between training and inference time.
We speculate that this will hinder the model's understanding of the relation between intrinsics, image shape, and metric depth.
On the contrary, our canonical image space will keep the image and intrinsic consistent.
As shown in the ablation studies of the main paper, the model benefits from the proposed canonical image space for both closed-set and open-set settings with even fewer GPU resource requirements for training. 

\begin{table}[t]
    \small
    \footnotesize
    \centering
    \caption{
        \textbf{GPU RAM Consumption}.
        We compare the GPU resource usage of different training padding and resizing methods.
        We show the results of training our full model using Swin-T~\cite{liu2021swin} and batch size of $2$ using gradient checkpointing on a RTX 4090.
    }
    \resizebox{\linewidth}{!}{
    \begin{tabular}{l|cc|c}
        \toprule
        Resize & Padding & Image Resolutions & GPU RAM (G) \\
        \midrule
        Short Edge & Right-Bottom & Short Edge to $512$ & 21 \\
        Long Edge & Right-Bottom & Long Edge to $1333$ & 23 \\
        Ours & Center & $800 \times 1333$ & 17 \\
    \bottomrule
    \end{tabular}}
    \label{tab:canonical}
\end{table}

\begin{table}[t]
    \small
    \footnotesize
    \centering
    \caption{
        \textbf{Comparison with Virtual Depth}.
        We compare our geometry-aware 3D query generation (GA) with the virtual depth proposed by Cube R-CNN~\cite{brazil2023omni3d}.
        The results shows that GA converge better than the virtual depth mechanism.
    }
    \resizebox{0.8\linewidth}{!}{
    \begin{tabular}{l|cc|c}
        \toprule
        Method & Virtual Depth & GA & $\text{AP}^{\text{ omni}}_{\text{3D}}  \uparrow$ \\
        \midrule
        Cube R-CNN~\cite{brazil2023omni3d} & \checkmark & - & 23.3 \\
        \midrule
        Ours (Swin-T) & \checkmark & - & 21.6 \\
        Ours (Swin-T) & - & \checkmark & \textbf{26.8} \\
    \bottomrule
    \end{tabular}}
    \label{tab:sup:virtual_depth}
\end{table}

\section{Comparison with Virtual Depth}
\label{sec:sup:virtual_depth}
We compare our proposed geometry-aware 3D query generation with the virtual depth proposed in~\cite{brazil2023omni3d}.
As shown in \cref{tab:sup:virtual_depth}, the virtual depth leads our model to converge much slower than our proposed geometry-aware 3D query generation.
We speculate that virtual depth requires much more training time to learn universal geometry, which also leads to underperformance.

\section{Open-set Benchmark}
\label{sec:sup:open}
We show more details about our proposed open-set benchmark and list the full classes for each datasets in~\cref{tab:sup:open_classes}.

\begin{table*}[ht]
    \centering
    \caption{
        \textbf{Classes for Argoverse 2 and ScanNet.}
        We list the base and novel categories for our proposed open-set benchmark.
        For ScanNet, the bold categories are the supercategories.
        We further list all $168$ categories of ScanNet200 settings.
    }
    \resizebox{\linewidth}{!}{
    \begin{tabular}{l|p{8cm}|p{8cm}}
        \toprule
        Dataset & Base & Novel \\
        \midrule
        Argoverse 2 & regular vehicle, pedestrian, bicyclist, construction cone, construction barrel, large vehicle, bus, truck, vehicular trailer, bicycle, motorcycle & motorcyclist, wheeled rider, bollard, sign, stop sign, box truck, articulated bus, mobile pedestrian crossing sign, truck cab, school bus, wheeled device, stroller \\
        \midrule
        ScanNet & \textbf{cabinet} (cabinet, kitchen cabinet, file cabinet, bathroom vanity, bathroom cabinet, cabinet door, trash cabinet, media center), \textbf{bed} (bed, mattress, loft bed, sofa bed, air mattress), \textbf{chair} (chair, office chair, armchair, sofa chair, folded chair, massage chair, recliner chair, rocking chair), \textbf{sofa} (couch, sofa), \textbf{table} (table, coffee table, end table, dining table, folded table, round table, side table, air hockey table), \textbf{door} (door, doorframe, bathroom stall door, closet door, mirror door, glass door, sliding door, closet doorframe), \textbf{window}, \textbf{picture} (picture, poster, painting), \textbf{counter} (kitchen counter, counter, bathroom counter), \textbf{desk}, \textbf{curtain}, \textbf{refrigerator} (refrigerator, mini fridge, cooler), \textbf{toilet} (toilet, urinal), \textbf{sink}, \textbf{bathtub} & \textbf{bookshelf}, \textbf{shower curtain}, \textbf{other furniture} (trash can, radiator, recycling bin, ottoman, bench, tv stand, wardrobe, trash bin, seat, closet, ladder, piano, water cooler, stand, washing machine, rack, wardrobe , clothes dryer, ironing board, keyboard piano, music stand, furniture, crate, drawer, footrest, piano bench, foosball table, footstool, compost bin, tripod, treadmill, chest, folded ladder, drying rack, pool table, heater, toolbox, beanbag chair, dollhouse, ping pong table, clothing rack, podium, luggage stand, rack stand, futon, book rack, workbench, easel, headboard, display rack, crib, bedframe, bunk bed, magazine rack, furnace, stepladder, baby changing station, flower stand, display) \\
        \midrule
        ScanNet200 & \multicolumn{2}{p{16.4cm}}{
            chair, table, door, couch, cabinet, shelf, desk, office chair, bed, pillow, sink, picture, window, toilet, bookshelf, monitor, curtain, book, armchair, coffee table, box, refrigerator, lamp, kitchen cabinet, towel, clothes, tv, nightstand, counter, dresser, stool, plant, bathtub, end table, dining table, keyboard, bag, backpack, toilet paper, printer, tv stand, whiteboard, blanket, shower curtain, trash can, closet, stairs, microwave, stove, shoe, computer tower, bottle, bin, ottoman, bench, board, washing machine, mirror, copier, basket, sofa chair, file cabinet, fan, laptop, shower, paper, person, paper towel dispenser, oven, blinds, rack, plate, blackboard, piano, suitcase, rail, radiator, recycling bin, container, wardrobe, soap dispenser, telephone, bucket, clock, stand, light, laundry basket, pipe, clothes dryer, guitar, toilet paper holder, seat, speaker, column, ladder, cup, jacket, storage bin, coffee maker, dishwasher, paper towel roll, machine, mat, windowsill, bar, bulletin board, ironing board, fireplace, soap dish, kitchen counter, doorframe, toilet paper dispenser, mini fridge, fire extinguisher, ball, hat, shower curtain rod, water cooler, paper cutter, tray, pillar, ledge, toaster oven, mouse, toilet seat cover dispenser, cart, scale, tissue box, light switch, crate, power outlet, decoration, sign, projector, closet door, vacuum cleaner, headphones, dish rack, broom, range hood, hair dryer, water bottle, vent, mailbox, bowl, paper bag, projector screen, divider, laundry detergent, bathroom counter, stick, bathroom vanity, closet wall, laundry hamper, bathroom stall door, ceiling light, trash bin, dumbbell, stair rail, tube, bathroom cabinet, coffee kettle, shower head, case of water bottles, power strip, calendar, poster, mattress
        } \\
        \bottomrule
    \end{tabular}
    }
    \label{tab:sup:open_classes}
\end{table*}

\subsection{Argoverse 2}
Compared to other autonomous driving datasets, Argoverse 2 (AV2)~\cite{Argoverse2} possesses more diverse classes.
Moreover, the resolutions of the front camera are portrait images, providing unseen cameras and domains.
Those factors make AV2 a challenging dataset for benchmarking open-set monocular 3D object detection.
We sample every $5$ frame from the official validation split and obtain $4806$ images as the open-set testing set.
Among the official $26$ classes, $23$ appeared in the testing set, which contains $11$ \textit{base} and $12$ \textit{novel} classes.

\subsection{ScanNet}
ScanNet~\cite{dai2017scannet} provides diverse indoor scenes with $18$ supercategories as shown in \cref{tab:sup:open_classes}.
We uniformly sample maximum $20$ frames from each scan in the official ScanNet validation splits and obtain total $6240$ images as open-set testing set.
Given that $15$ supercategories are seen in the Omni3D training set, this benchmark still allows us to evaluate domain generalization, where Tab.~4 of the main paper indicates issues of previous works.
Furthermore, in the rest $3$ novel classes, the supercategory \textit{other furniture} requires models to detect various types of furniture.

To further test \ourmodel, we extend ScanNet using the ScanNet200 setting, which has $\textbf{168}$ thing classes appeared in the testing set.
As shown in~\cref{tab:sup:scannet200}, \ourmodel can still achieve best performance given more diverse classes.

\begin{table}[t]
    \small
    \footnotesize
    \centering
    \caption{
        \textbf{ScanNet200 Results.}
        \ourmodel achieves SOTA results given diverse novel categories in unseen scenes.
    }
    \resizebox{\linewidth}{!}{
    \begin{tabular}{l|ccccc}
        \toprule
        Method & $\text{AP}_{\text{3D}}^{\text{ dist}} \uparrow$ & mATE $\downarrow$ & mASE $\downarrow$ & mAOE $\downarrow$ & \ourmetric $\uparrow$ \\
        \midrule
        Cube R-CNN~\cite{brazil2023omni3d} & 2.1 & 0.962 & 0.970 & 0.985 & 2.5 \\
        OVM3D-Det~\cite{huang2024training} & 3.1 & 0.957 & 0.973 & 0.946 & 3.6 \\
        \textbf{Ours} (Swin-B) & \textbf{6.2} & \textbf{0.811} & \textbf{0.835} & \textbf{0.799} & \textbf{12.4} \\
        \bottomrule
    \end{tabular}
    }
    \label{tab:sup:scannet200}
\end{table}

\section{Metric Monocular Depth Estimation}
\label{sec:sup:mmde}
Because auxiliary depth estimation (ADE) is only used to help with 3D object detection, we evaluated its effectiveness in this regard.
As shown in Tab.~4 of our paper, ADE improves the closed set AP by $0.7$, but reduces the performance for unseen scenes, indicating that ADE can only help for known scenes.
We further evaluate our depth quality on the KITTI Eigen-split test set, where UniDepth~\cite{piccinelli2024unidepth} achieves $4.21\%$ absolute relative error, Metric3Dv2~\cite{yin2023metric3d} has $4.4\%$, and \ourmodel obtains $9.1\%$.
We believe it is due to limited training data, different training objectives, and the model backbones~\cite{liu2021swin, oquab2023dinov2}.

\section{Backbone Comparison}
\label{sec:sup:backbone}
As shown in~\cref{tab:sup:backbone_ablation}, Swin-B works equally well with ConvNeXt-B, and \ourmodel is comparable to Cube R-CNN and Uni-MODE using the same backbone, but with much shorter training.
This shows effectiveness of our proposed designs rather than the backbone~\cite{oquab2023dinov2}.

\begin{table}[t]
    \small
    \footnotesize
    \centering
    \caption{
        \textbf{Backbone comparison.}
        We ablate the choice of different model backbones.
        All experiments are trained with $12$ epochs.
    }
    \resizebox{\linewidth}{!}{
    \begin{tabular}{l|c|c}
        \toprule
        Backbone & Parameters & $\text{AP}^{\text{ omni}}_{\text{3D}} \uparrow $ \\
        \midrule
        DLA-34~\cite{yu2018deep} & 15M & 24.9 \\
        Swin-Transformer (Tiny)~\cite{liu2021swin} & 29M & 26.8 \\
        Swin-Transformer (Base)~\cite{liu2021swin} & 88M & 28.2 \\
        ConvNeXt-B~\cite{liu2022convnet} & 89M & 28.4 \\
        \bottomrule
    \end{tabular}
    }
    \label{tab:sup:backbone_ablation}
\end{table}

\begin{table}[t]
    \small
    \footnotesize
    \centering
    \caption{
        \textbf{Comparison between different matching criteria for evaluation.}
        The same detection results will have a huge AP difference when using different matching settings.
    }
    \resizebox{\linewidth}{!}{
    \begin{tabular}{l|cc|cc}
        \toprule
        Matching & Pedestrian & Construction cone & Monitor & Door \\
        \midrule
        IoU & 7.4 & 0.5 & 0.5 & 2.0 \\ 
        Distance & 26.2 & 6.5 & 9.4 & 24.2 \\
        \bottomrule
    \end{tabular}
    }
    \label{tab:sup:dist}
\end{table}

\section{Inference Time}
\label{sec:sup:fps}
We compare the FPS on KITTI using an RTX 4090, and Cube R-CNN (DLA-34) can have $68$ FPS while \ourmodel (Swin-T) can achieve $17$ FPS.
As a reference from the paper, Uni-MODE can obtain 21 FPS on a single A100.

\section{Open Detection Score (ODS)}
\label{sec:sup:eval}
Compared to point-cloud-based 3D object detectors, using a single image to estimate 3D objects requires the networks to predict metric depth, while the scales in depth are known in the point cloud.
This extra challenge leads the monocular methods to fail to match the ground truth using IoU-based matching because of several centimeter error in depth, especially for the small or thin objects in the open-set settings.

To have a more suitable evaluation metric for monocular 3D object detection, we use the 3D Euclidean distance between prediction and GT as the matching criterion.
With the dynamic matching threshold, \eg \textit{radius} of the GT 3D boxes, $\text{AP}_\text{3D}^\text{ dist}$ can be used for both indoor and outdoor scenes.
As shown in~\cref{tab:sup:dist}, the same detection results on Argoverse 2~\cite{Argoverse2} and ScanNet~\cite{dai2017scannet} will have large AP differences depending on the matching criterion.

We also propose the normalized true positive errors (TPE) to further analyze the matched prediction.
First, we compute the 3D Euclidean distance between prediction and GT, and we normalize the distance by the matching criterion as the translation error (TE).
Second, we compute the IoU, \ie $\text{IoU}_{\text{3D}}$, between prediction and GT after aligning the 3D centers and orientation and use $1 - \text{IoU}_{\text{3D}}$ to measure the scale error (SE).
Finally, we compute the SO3 relative angle between the prediction and GT normalized by $\pi$ as the orientation error (OE).
We average the TP errors across classes over different recall thresholds to get mATE, mASE, and mAOE.
Using $\text{AP}_\text{3D}^\text{ dist}$ with the proposed normalized true positive errors to get \ourmetric can provide a better matching criterion for 3D monocular object detection and still evaluate the localization, orientation, and dimension estimation at the same time.



\begin{figure}[t]
    \footnotesize
    \small
    \centering
    \setlength{\tabcolsep}{2pt}
    \begin{tabular}{cc}
        \includegraphics[width=0.45\linewidth, height=2.8cm]{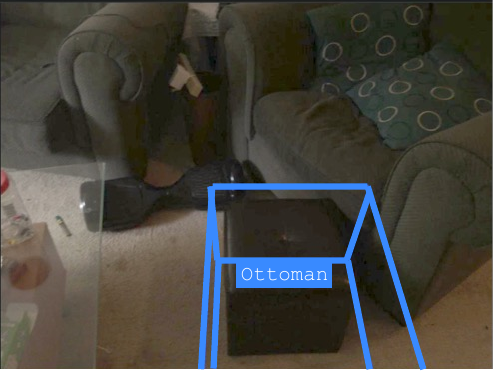} &
        \includegraphics[width=0.45\linewidth, height=2.8cm]{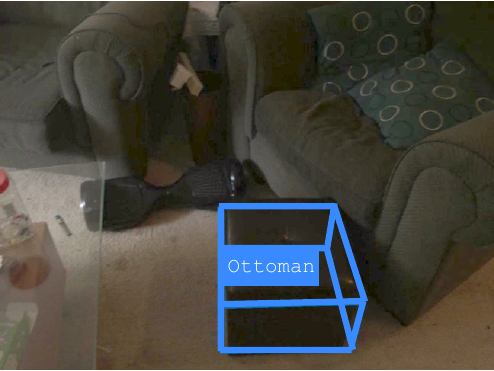} \\
        \includegraphics[width=0.45\linewidth, height=2.8cm]{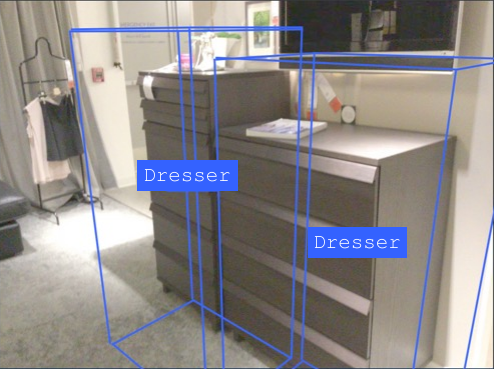} &
        \includegraphics[width=0.45\linewidth, height=2.8cm]{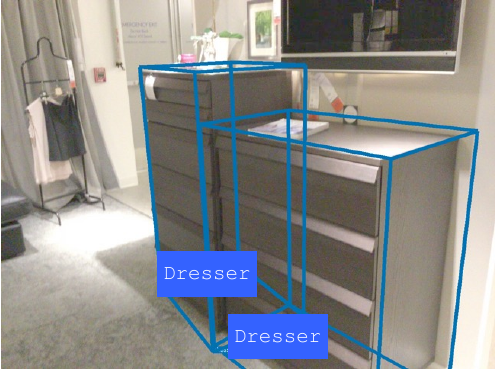} \\
        Gemini 2 & \textbf{\ourmodel} \\
    \end{tabular}
    \caption{
        \textbf{Comparison with Gemini 2.}
        We qualitatively compare with Gemini 2 given the novel classes.
    }
    \label{fig:gemini}
\end{figure}

\section{Qualitative Results}
\label{sec:sup:qualitative}
\begin{figure*}[t]
    \centering
    \small
    \footnotesize
    \setlength{\tabcolsep}{5pt}
    \newcommand{\sz}{0.22}
    \newcommand{\hs}{2.6cm}
    \begin{tabular}{ccccc}
        \\
        \rotatebox{90}{\hspace{45pt} Argoverse 2} &
        \includegraphics[width=\sz\linewidth, height=5cm]{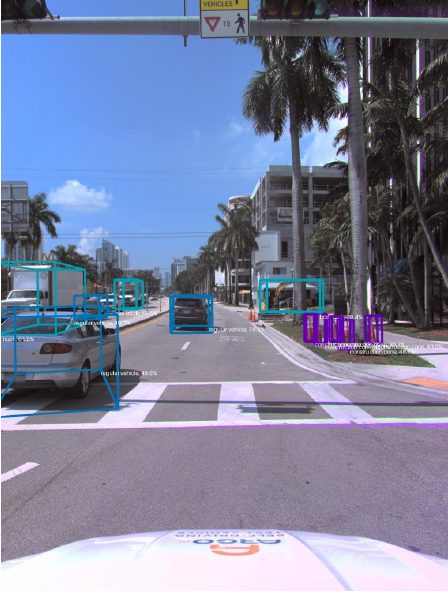} &
        \includegraphics[width=\sz\linewidth, height=5cm]{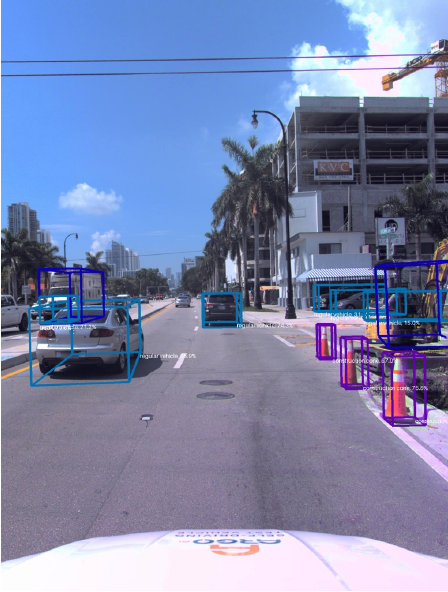} &
        \includegraphics[width=\sz\linewidth, height=5cm]{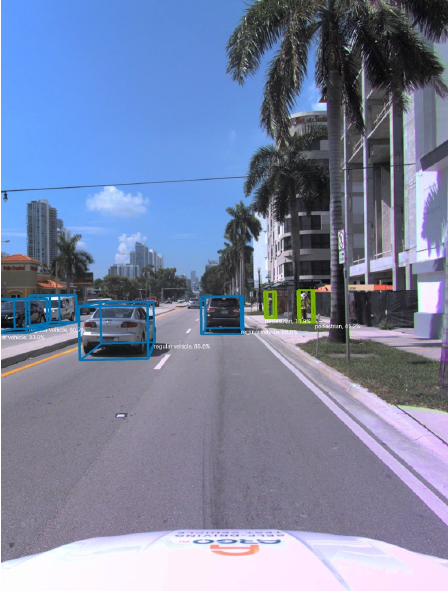} &
        \includegraphics[width=\sz\linewidth, height=5cm]{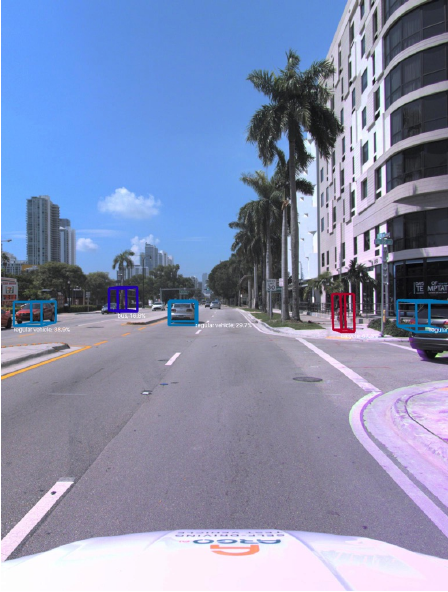} \vspace{5pt} \\
        \rotatebox{90}{\hspace{45pt} Argoverse 2} &
        \includegraphics[width=\sz\linewidth, height=5cm]{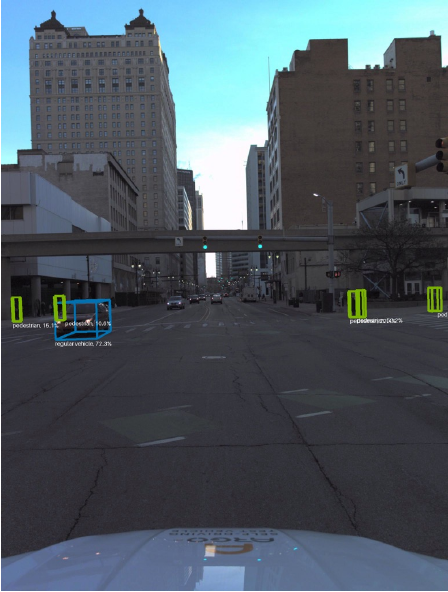} &
        \includegraphics[width=\sz\linewidth, height=5cm]{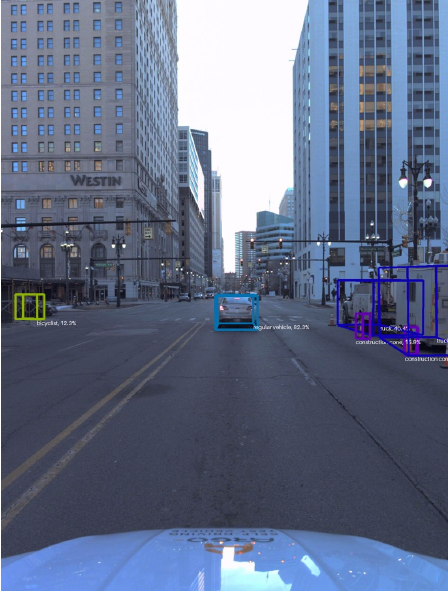} &
        \includegraphics[width=\sz\linewidth, height=5cm]{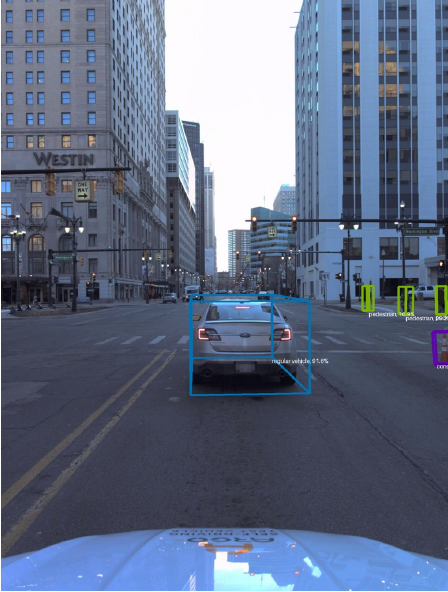} &
        \includegraphics[width=\sz\linewidth, height=5cm]{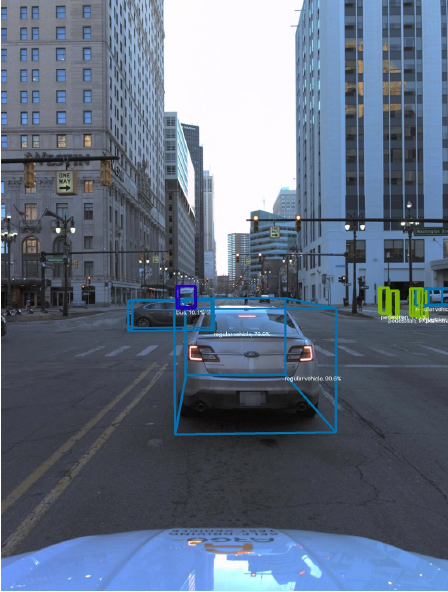} \vspace{5pt} \\
        \rotatebox{90}{\hspace{20pt} ScanNet}
        &
        \includegraphics[width=\sz\linewidth, height=\hs]{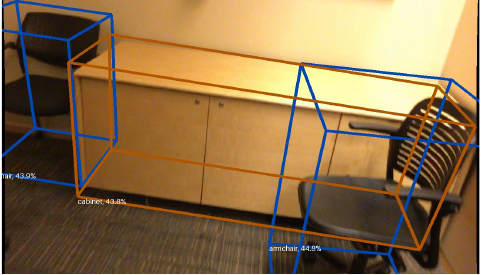} &
        \includegraphics[width=\sz\linewidth, height=\hs]{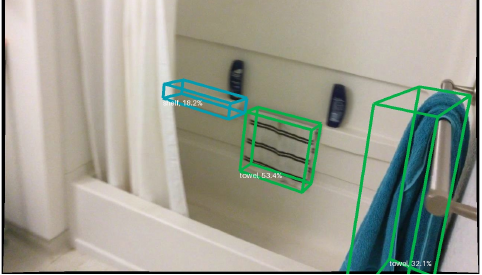} &
        \includegraphics[width=\sz\linewidth, height=\hs]{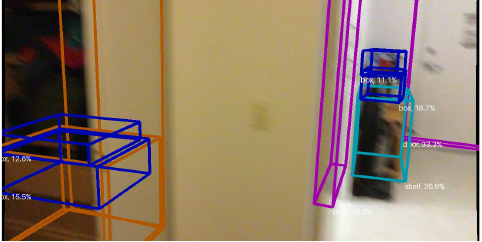} &
        \includegraphics[width=\sz\linewidth, height=\hs]{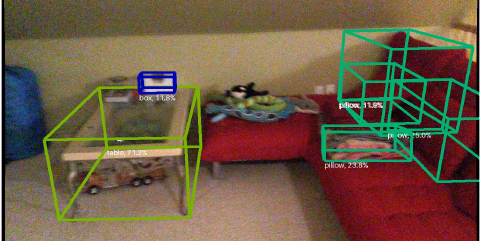} \vspace{5pt} \\ 
        \rotatebox{90}{\hspace{20pt} ScanNet}
        &
        \includegraphics[width=\sz\linewidth, height=\hs]{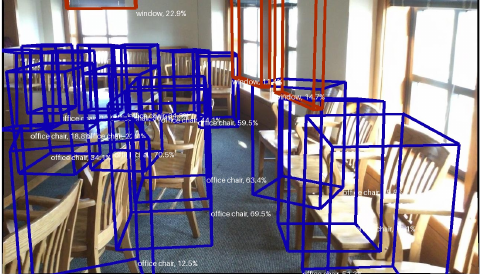} &
        \includegraphics[width=\sz\linewidth, height=\hs]{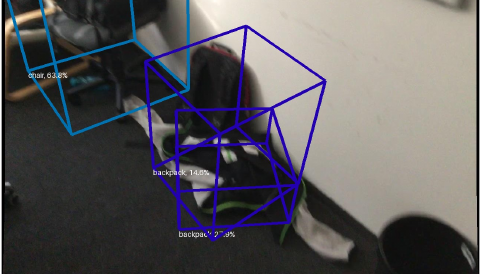} &
        \includegraphics[width=\sz\linewidth, height=\hs]{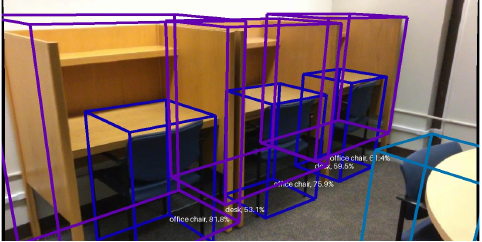} &
        \includegraphics[width=\sz\linewidth, height=\hs]{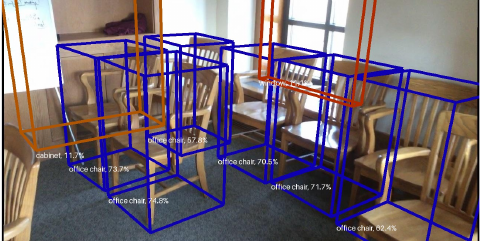} \vspace{5pt} \\ 
        \rotatebox{90}{\hspace{24pt} 3D} & \includegraphics[width=\sz\linewidth]{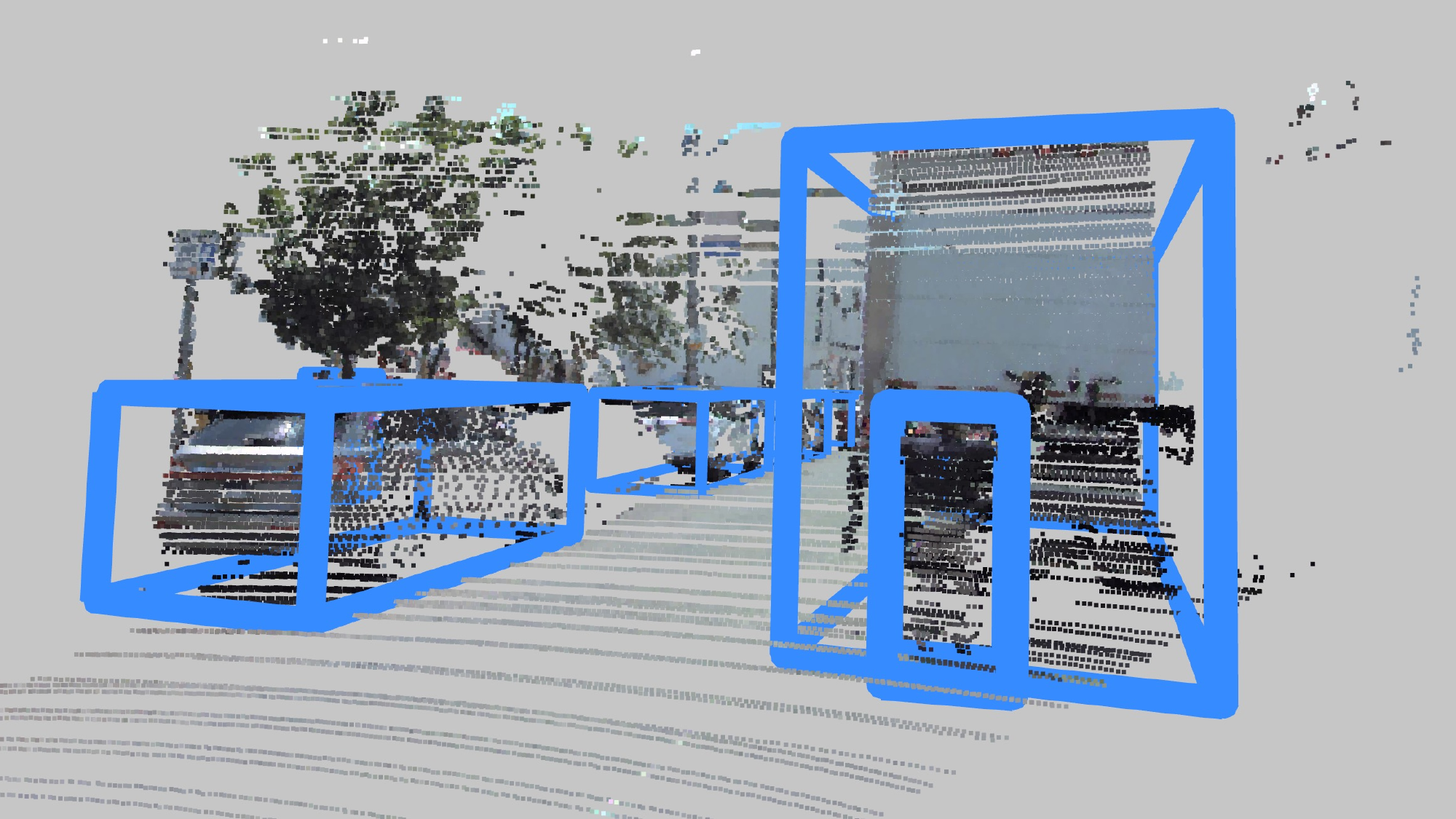} & \includegraphics[width=\sz\linewidth]{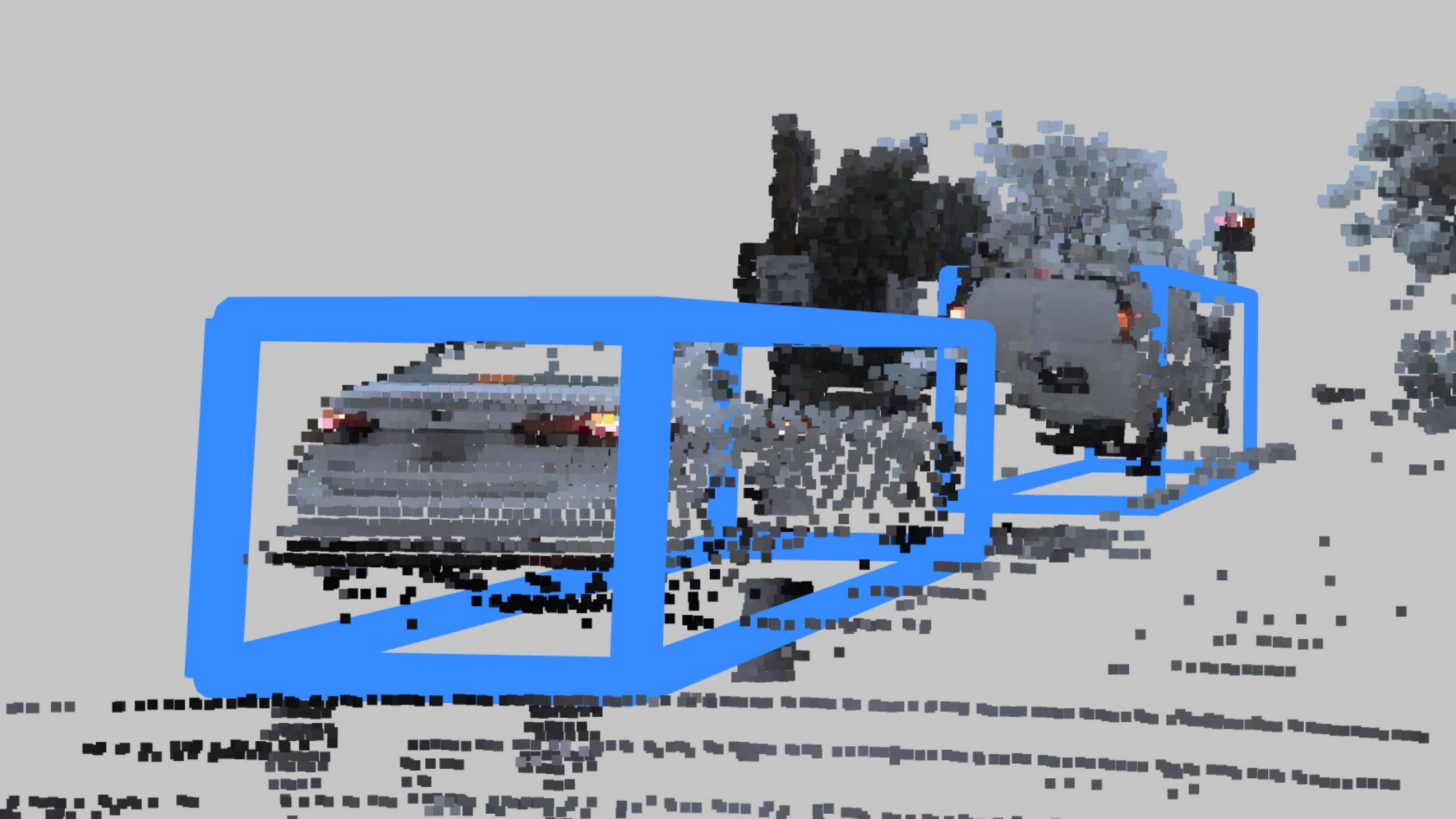} & \includegraphics[width=\sz\linewidth]{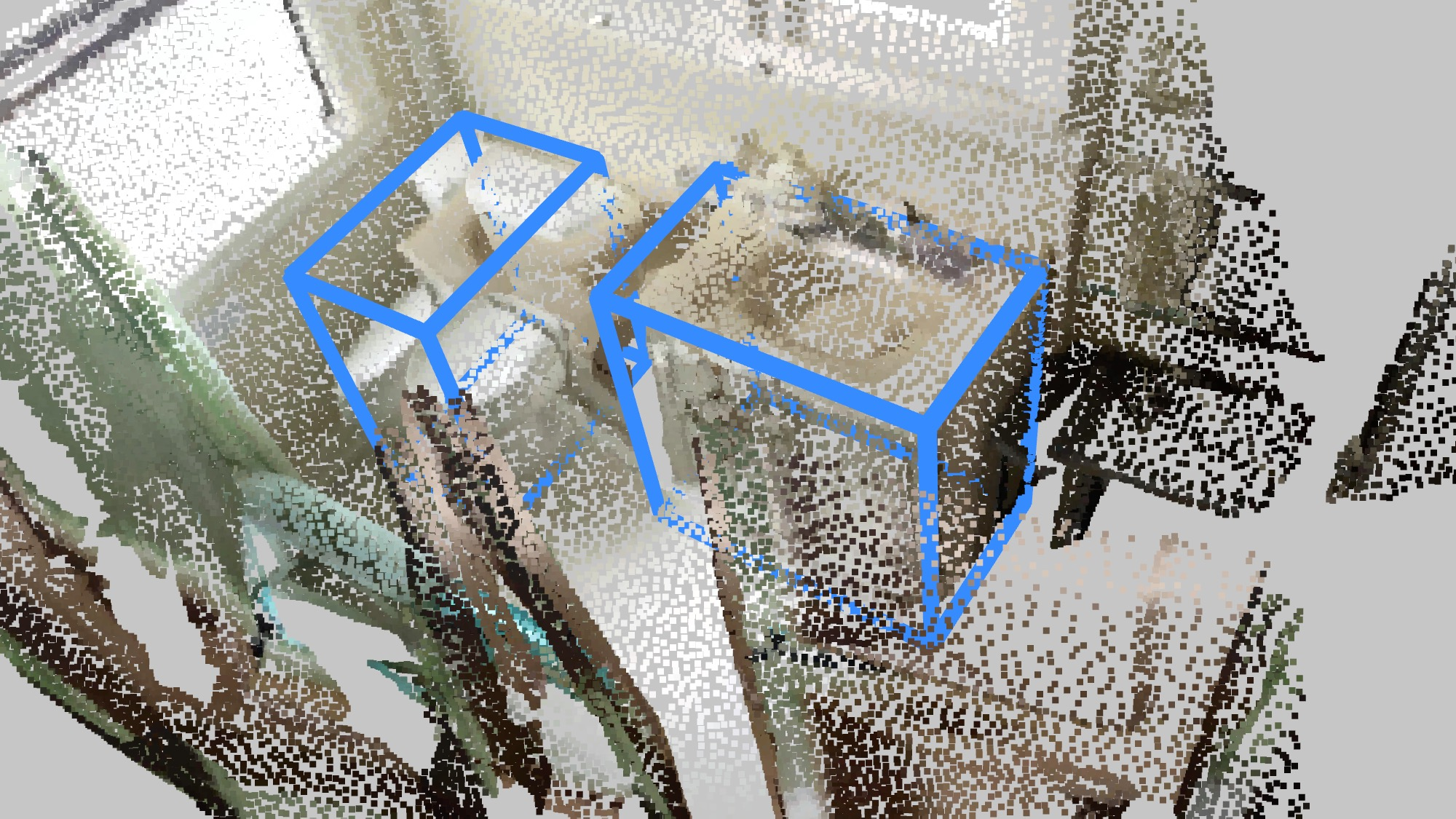} & \includegraphics[width=\sz\linewidth]{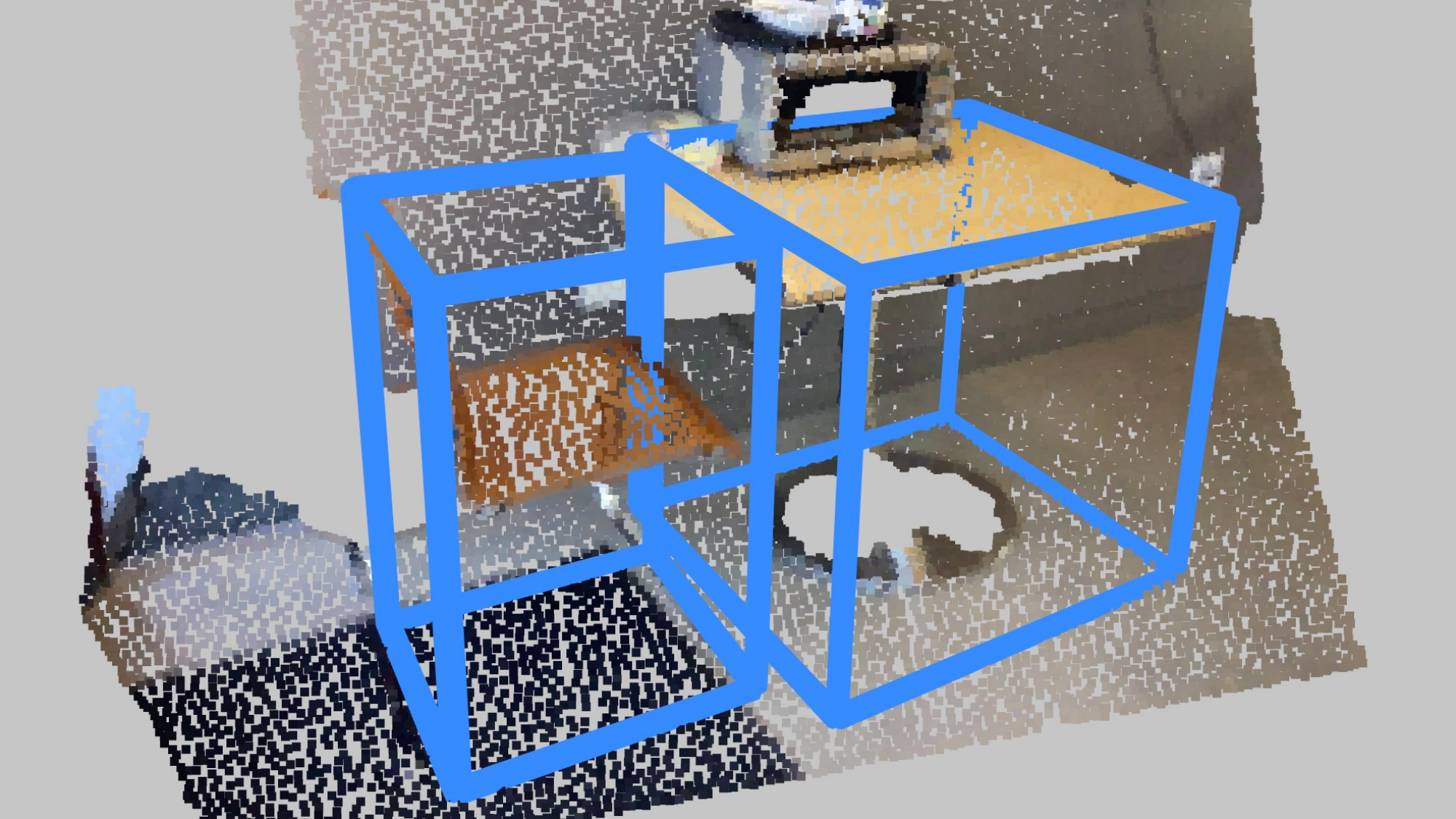} \\
    \end{tabular}
    \caption{
        \textbf{Open-set Qualitative Results.}
        We show more visualization on Argoverse 2~\cite{Argoverse2} and ScanNet~\cite{dai2017scannet}.
    }
    \label{fig:sup:open_qualitative}
\end{figure*}

\begin{figure*}[t]
    \centering
    \small
    \setlength{\tabcolsep}{2pt}
    \newcommand{\sz}{0.22}
    \begin{tabular}{ccccc}
        \rotatebox{90}{\hspace{3pt} KITTI} &
        \includegraphics[width=\sz\linewidth]{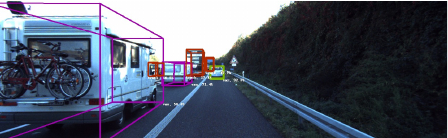} &
        \includegraphics[width=\sz\linewidth]{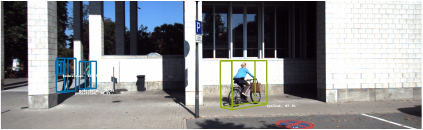} &
        \includegraphics[width=\sz\linewidth]{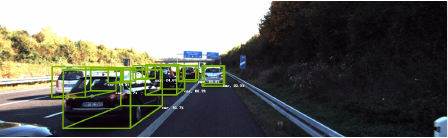} &
        \includegraphics[width=\sz\linewidth]{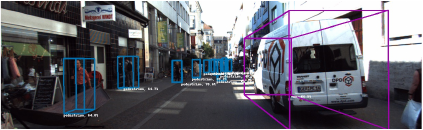} \vspace{3pt} \\
        \rotatebox{90}{\hspace{10pt} nuScenes} &
        \includegraphics[width=\sz\linewidth]{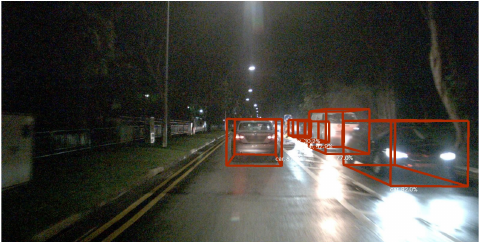} &
        \includegraphics[width=\sz\linewidth]{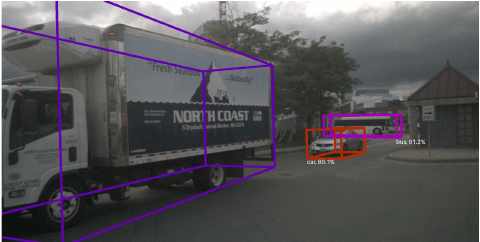} &
        \includegraphics[width=\sz\linewidth]{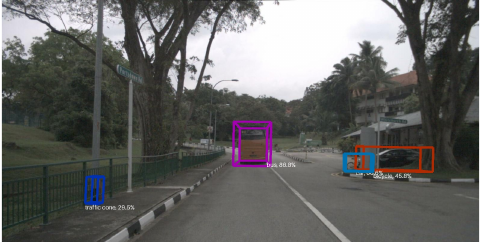} &
        \includegraphics[width=\sz\linewidth]{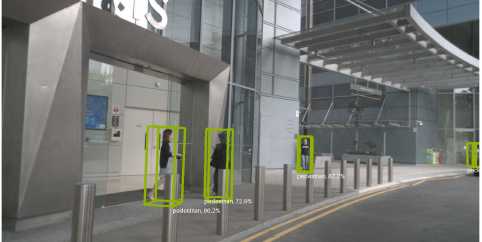} \vspace{3pt} \\
        \rotatebox{90}{\hspace{32pt} Objectron} &
        \includegraphics[width=\sz\linewidth, height=3.6cm]{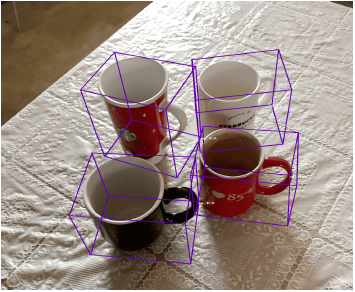} &
        \includegraphics[width=\sz\linewidth, height=3.6cm]{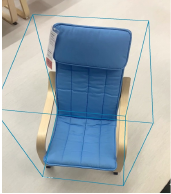} &
        \includegraphics[width=\sz\linewidth, height=3.6cm]{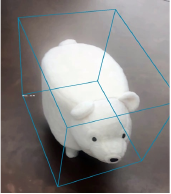} &
        \includegraphics[width=\sz\linewidth, height=3.6cm]{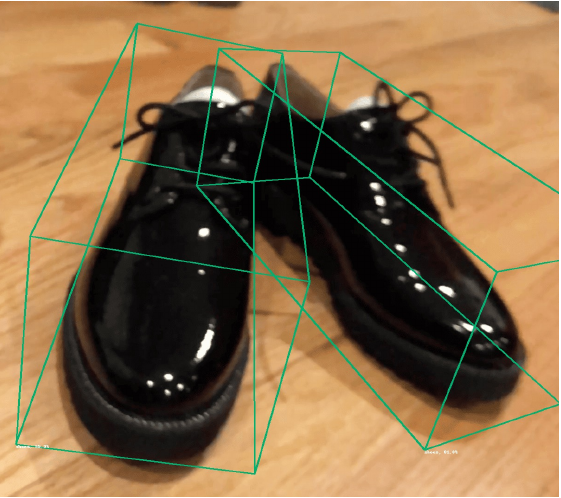} \vspace{3pt} \\
        \rotatebox{90}{\hspace{30pt} Hypersim} &
        \includegraphics[width=\sz\linewidth, height=3.2cm]{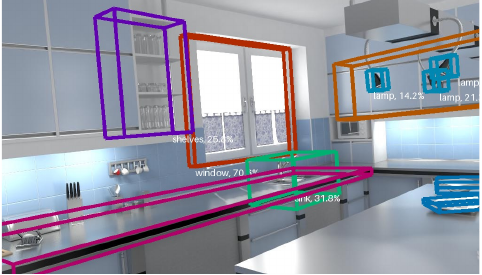} &
        \includegraphics[width=\sz\linewidth, height=3.2cm]{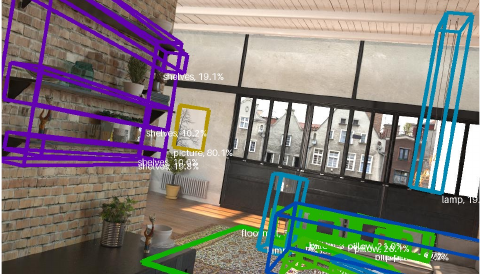} &
        \includegraphics[width=\sz\linewidth, height=3.2cm]{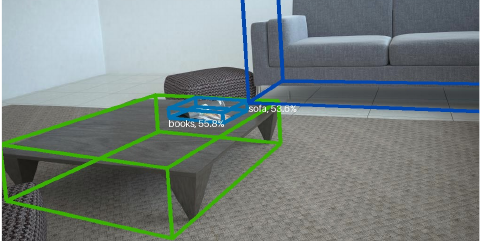} &
        \includegraphics[width=\sz\linewidth, height=3.2cm]{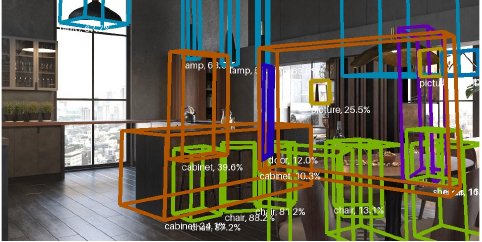} \vspace{3pt} \\
        \rotatebox{90}{\hspace{20pt} ARKitScenes} &
        \includegraphics[width=\sz\linewidth, height=3.2cm]{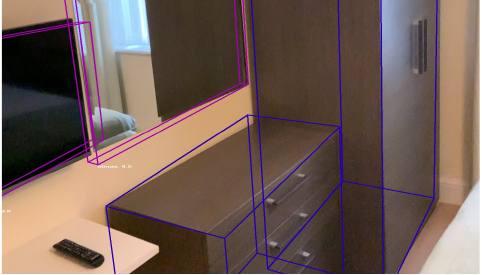} &
        \includegraphics[width=\sz\linewidth, height=3.2cm]{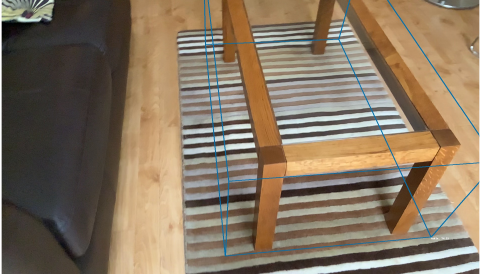} &
        \includegraphics[width=\sz\linewidth, height=3.2cm]{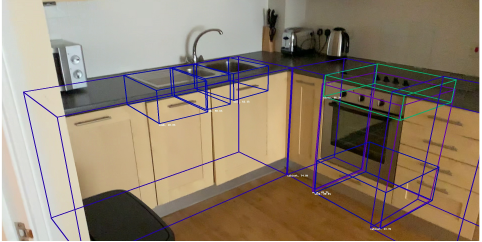} &
        \includegraphics[width=\sz\linewidth, height=3.2cm]{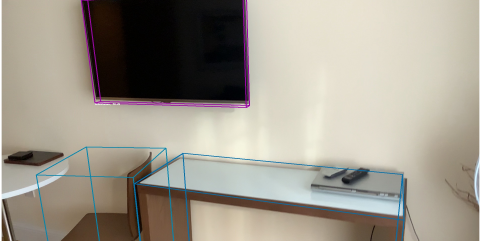} \vspace{3pt} \\
        \rotatebox{90}{\hspace{27pt} SUN RGB-D} &
        \includegraphics[width=\sz\linewidth, height=3.4cm]{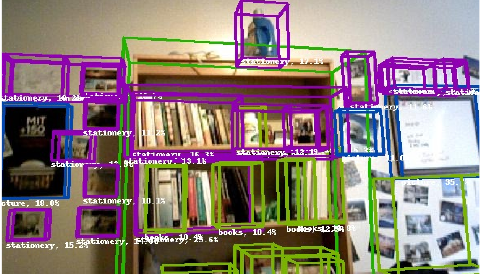} &
        \includegraphics[width=\sz\linewidth, height=3.4cm]{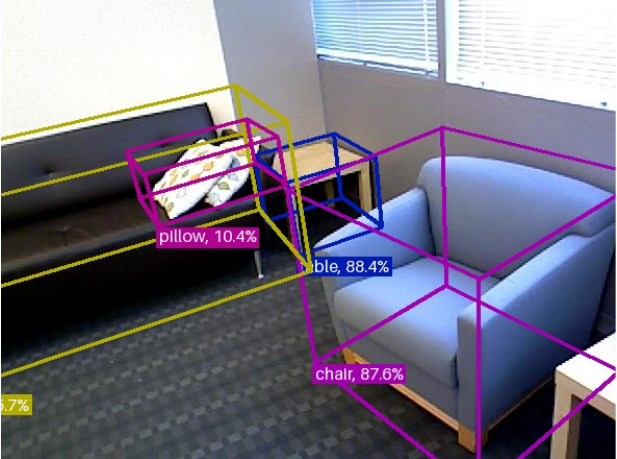} &
        \includegraphics[width=\sz\linewidth, height=3.4cm]{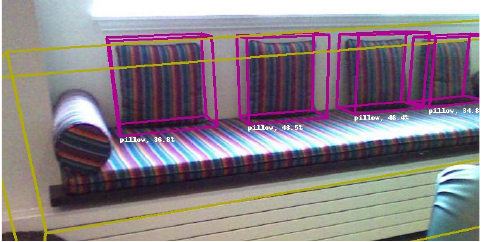} &
        \includegraphics[width=\sz\linewidth, height=3.4cm]{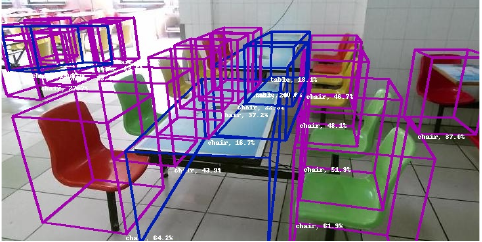} \vspace{3pt} \\
        \rotatebox{90}{\hspace{25pt} 3D} & \includegraphics[width=\sz\linewidth]{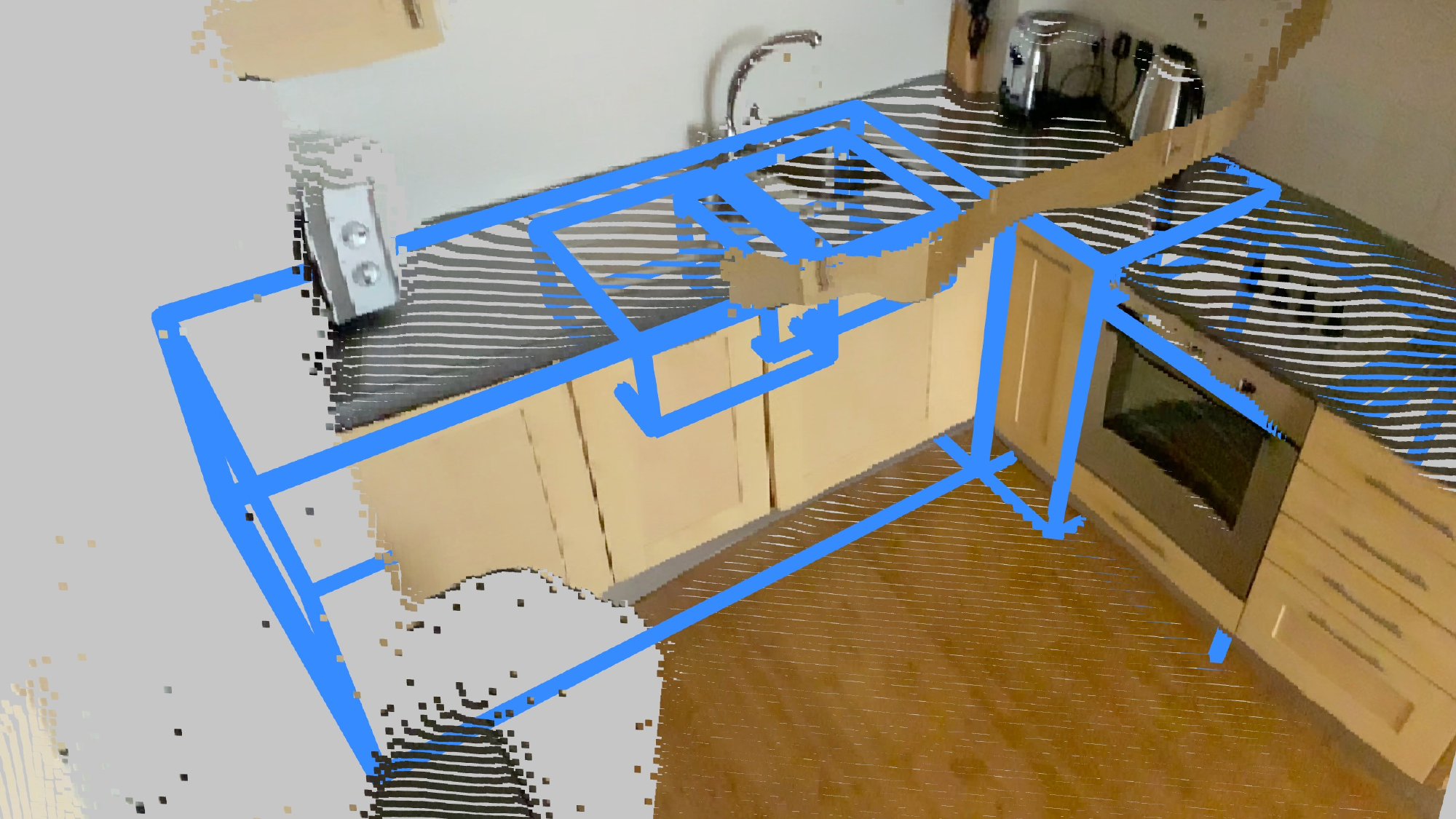} & \includegraphics[width=\sz\linewidth]{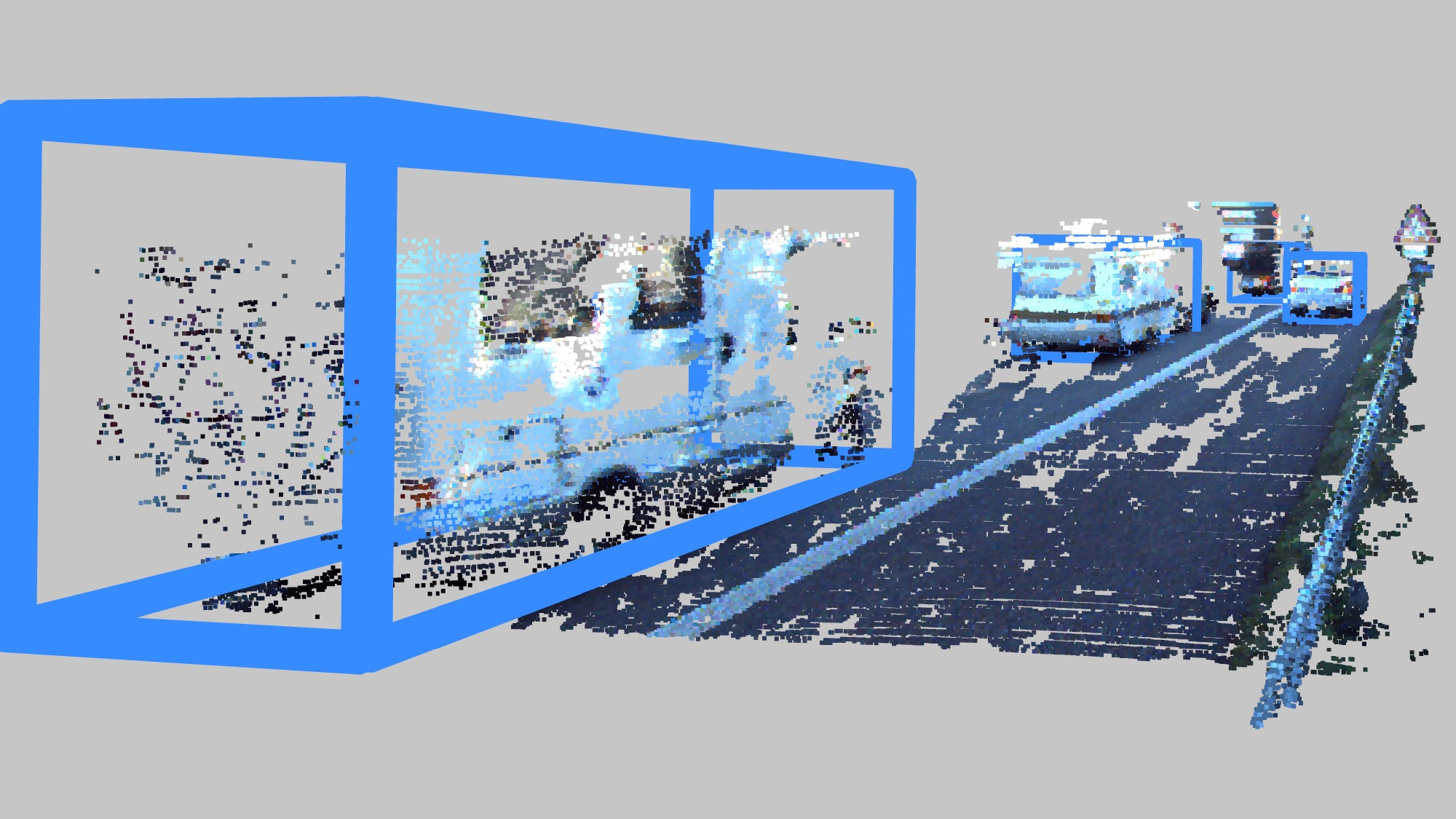} & \includegraphics[width=\sz\linewidth]{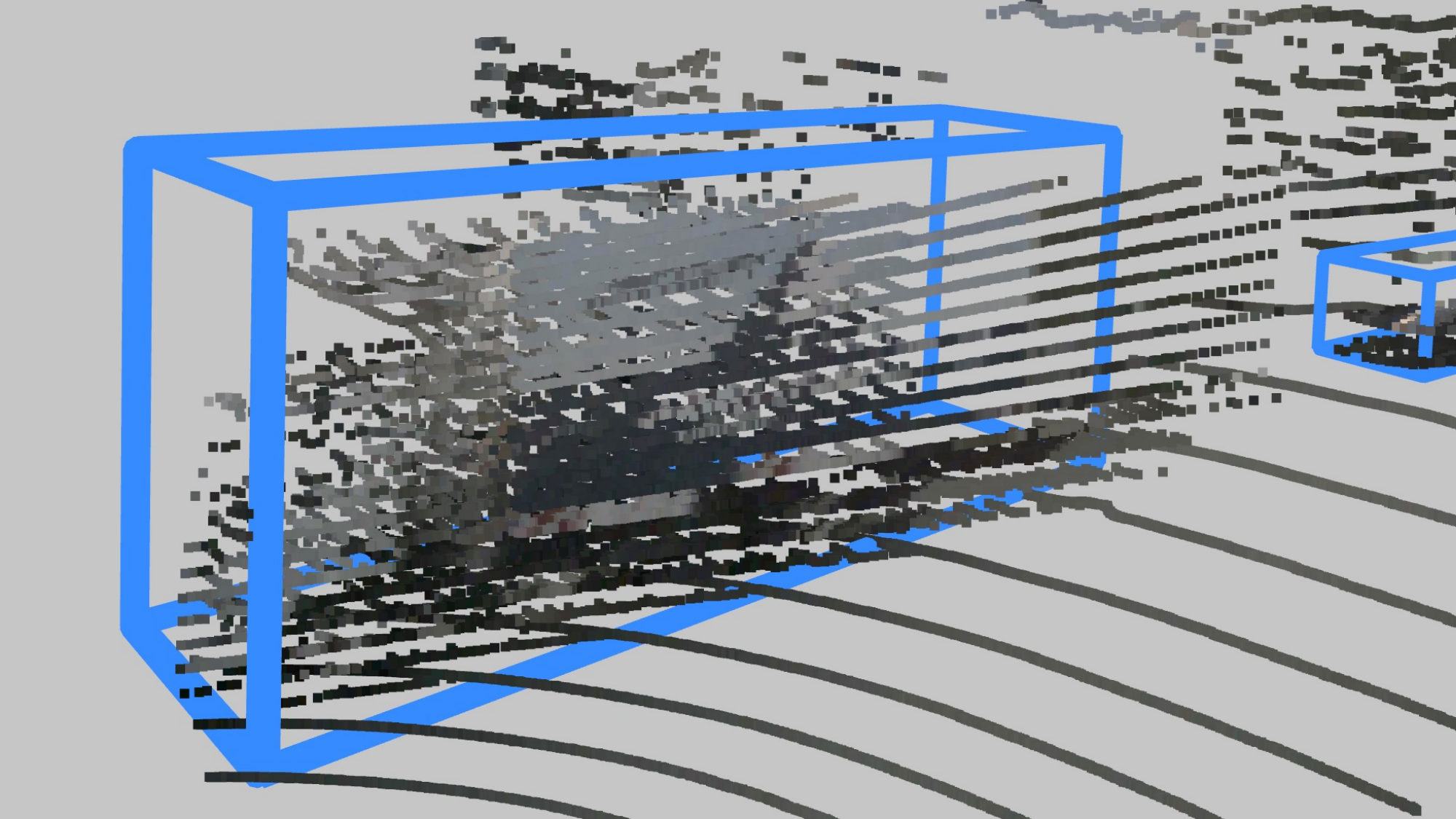} & \includegraphics[width=\sz\linewidth]{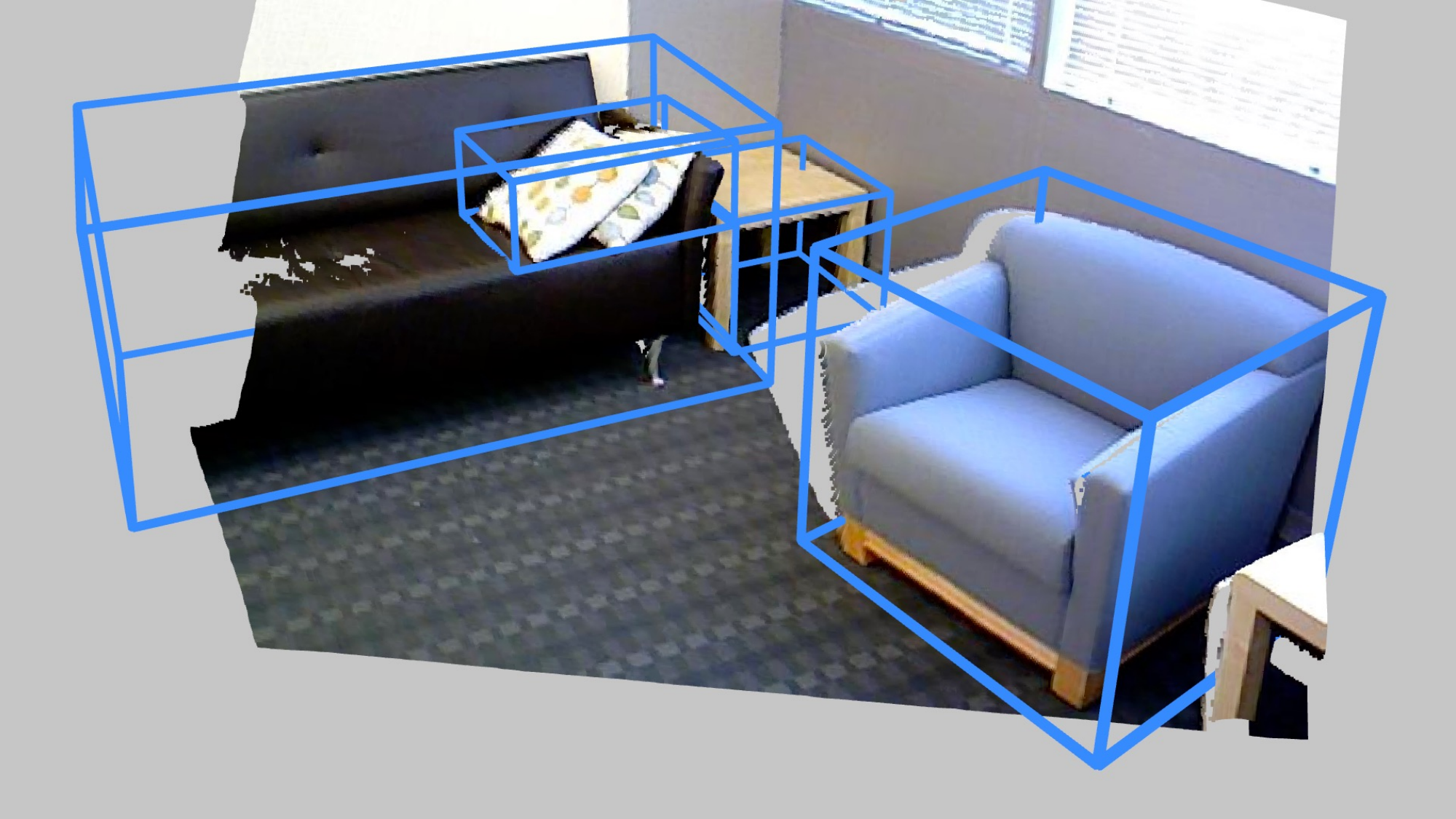} \\
    \end{tabular}
    \vspace{-5pt}
    \caption{
        \textbf{Closed-set Qualitative Results.}
        We show the qualitative results for \ourmodel on Omni3D~\cite{brazil2023omni3d} test set.
    }
    \label{fig:sup:omni3d_qualitative}
\end{figure*}

As shown in~\cref{fig:gemini}, we qualitatively compared our method with the closed-source Gemini 2~\cite{team2023gemini} beta functionality in 3D object detection, where \ourmodel provides more accurate localization.
We provide more qualitative results in~\cref{fig:sup:open_qualitative} for the open-set settings and~\cref{fig:sup:omni3d_qualitative} for the closed-set settings.
We use the score threshold as $0.1$ with class-agnostic nonmaximum suppression for better visualization.

\end{document}